%% file: main.tex
\pdfoutput=1
\documentclass[11pt]{article}
\input{preamble.tex}
\hypersetup{pdftitle={Cross-Person Identity Change Under Source-Coupled Multi-Human Generation: A Controlled UniPortrait Study},pdfauthor={Edrees Al-Edreesi, Ghaleb Al-Gaphari, Muneer ALSURORI, Mohammed Edrees, Fouad Al-Hamdani},pdfsubject={Controlled empirical study of person-wise identity similarity in UniPortrait}}
\title{Cross-Person Identity Change Under Source-Coupled Multi-Human Generation:\\A Controlled UniPortrait Study}
\author{%
  \parbox{\dimexpr\textwidth-2\tabcolsep\relax}{%
    \centering
    Edrees Al-Edreesi\textsuperscript{1,2,*},\quad
    Ghaleb Al-Gaphari\textsuperscript{1},\quad
    Muneer ALSURORI\textsuperscript{2},\quad
    Mohammed Edrees\textsuperscript{2},\quad
    Fouad Al-Hamdani\textsuperscript{2}\\[0.9em]
    \textsuperscript{1}\small Sana’a University, Sana’a, Yemen.\\[0.25em]
    \textsuperscript{2}\small Ibb University, Ibb, Yemen.\\[0.4em]
    \textsuperscript{*}\small Corresponding author: \texttt{adris.shrf@su.edu.ye}\\[0.3em]
    \small Project repository: \url{https://github.com/edrees2022/face-identity-repair}%
  }%
}
\date{}
\begin{document}
\maketitle
\begin{abstract}
\input{sections/abstract}
\end{abstract}
\input{sections/introduction}
\input{sections/related_work}
\input{sections/design_measurement}

\input{sections/empirical_findings}
\input{sections/validation}

\input{sections/discussion}
\input{sections/conclusion}
\FloatBarrier
\bibliographystyle{plainnat}
\bibliography{bibliography/references}
\end{document}


\maketitle
\noindent This Supplement provides detailed methods, statistical models, numerical results, additional figures, and ethics and data-availability disclosures for the two sequential studies and their validation analyses.
\setcounter{tocdepth}{1}
\tableofcontents
\clearpage
\input{supplement/methods}
\input{supplement/models}
\input{supplement/frozen_tables}
\input{supplement/independent_validation}
\input{supplement/nuisance}

\input{supplement/plots}
\input{supplement/provenance_ethics}
\FloatBarrier
\bibliographystyle{plainnat}
\bibliography{bibliography/references}

%% file: preamble.tex
\usepackage[T1]{fontenc}
\usepackage[utf8]{inputenc}
\usepackage{lmodern}
\usepackage[letterpaper,margin=0.8in]{geometry}
\usepackage{microtype,amsmath,amssymb,graphicx,booktabs,array,tabularx}
\usepackage{caption,subcaption,float,placeins,adjustbox,pdflscape}
\usepackage{tikz}
\usetikzlibrary{arrows.meta,positioning,calc}
\usepackage[numbers,sort&compress]{natbib}
\usepackage{xurl}
\usepackage[hidelinks]{hyperref}
\usepackage[nameinlink,noabbrev]{cleveref}
\usepackage{fancyhdr}
\newcolumntype{Y}{>{\raggedright\arraybackslash}X}
\newcommand{\ID}{\mathrm{ID}}
\newcommand{\dID}{\Delta\mathrm{ID}}
\newcommand{\PRE}{\mathrm{PRE}}
\newcommand{\POST}{\mathrm{POST}}
\newcommand{\Aone}{\mathrm{A1}}
\newcommand{\Atwo}{\mathrm{A2}}

\newcommand{\buffalo}{Buffalo\_L}
\graphicspath{{assets/png/}{assets/pdf/}}

%% file: sections/abstract.tex
Multi-human personalization must account for identity changes in people outside the nominal target of a generative operation. We characterize this behavior in UniPortrait, without introducing a new preservation algorithm. Two sequential studies use 119 accepted two-person baseline scenes and three fixed output seeds, measuring continuous reference-based facial similarity with InsightFace \buffalo. In Study~1, source-coupled single-ID target-only generation (A1) produces negative mean transitions for both the protected person ($-0.567904$) and target ($-0.236904$), indicating broad perturbation rather than established successful repair with collateral loss. In Study~2, native dual-ID generation (A2) yields more positive paired contrasts: protected $+0.512918$ (95\% case-cluster CI $[0.471434,0.550259]$) and target $+0.158853$ ($[0.118903,0.197523]$), across 287 jointly measurable pairs from 111 cases. Shared baseline similarity cancels algebraically in these paired contrasts, leaving a matched POST comparison. Validation designed after the primary result freeze finds directional agreement with an independent identity evaluator and blinded human judgments; within-\buffalo\ nuisance models provide supportive sensitivity evidence. The findings characterize intact native regimes, whose references, routing, adapters, and token counts co-vary. They do not establish component causality, categorical identity preservation, successful localized editing, or generalization beyond the evaluated implementation and measurable cohort.

%% file: sections/introduction.tex
\section{Introduction}
\label{sec:intro}
Preserving several identities in one generated scene is an established objective of image personalization. Multi-concept and multi-subject methods already address reference conditioning, subject separation, and identity interference \citep{kumari2023multi,xiao2023fastcomposer,jang2024mudi}. UniPortrait provides native single- and multi-human identity conditioning within one framework \citep{ma2024uniportrait}. Group portrait editing and facial restoration also explicitly address preservation of people affected by an edit \citep{groupdiff2024,editedid2024}. The scientific question here is narrower: how does a co-present person's measured facial similarity change when an existing implementation executes a target-directed conditioning regime, and how does that behavior differ across its intact native regimes?

This question matters because a nominal target is only one participant in a shared generative canvas. A person outside the target can undergo a measurable change even when their identity is not explicitly being steered. Conversely, an operation described as target-directed need not successfully improve the target. Evaluating both people against their own reference portraits separates these possibilities. It also avoids equating a better relative outcome with complete preservation. Our target and protected labels designate prospective evaluation roles; neither label establishes initial biometric adequacy, a failed target identity, or a person whose identity has already been verified as preserved.

We study two-person synthetic scenes from MultiHuman-Testbench \citep{mhtb2025}, using UniPortrait with a Stable Diffusion~1.5 backbone. Each accepted baseline image, called PRE, has a designated target $k$ and a different protected person $j$. Similarity to each person's fixed reference is measured before and after generation. The resulting continuous transition, $\dID_p$, records the change for role $p$. Negative values indicate lower reference similarity in the measured output. They are not probabilities of identity failure, and no threshold converts them into categorical judgments about loss, harm, or task success.

Study~1 characterizes source-coupled native single-ID target-only generation, A1, against an uncoupled global regeneration comparator, A0. Its central finding is negative mean similarity change for both people. The protected transition is pronounced, but the target transition is also negative. The dominant pattern therefore supports broad multi-person perturbation under this implementation. It does not support a headline in which successful target repair has a collateral identity cost. The comparator supplies context for the size of the observed change; it is not a localized editing baseline with the same source coupling or the same correspondence procedure.

Study~2 then compares frozen A1 outputs with prospectively generated native dual-ID outputs, A2, using the same PRE source trajectory and matched output seeds. Both protected and target contrasts are more positive under A2 among jointly measurable pairs. This comparison estimates the difference between complete operational configurations. It changes reference availability together with native routing, adapter weights, and identity-token count, so it does not isolate a mechanism. Moreover, the shared PRE similarity cancels in the paired A2$-$A1 difference. PRE remains necessary to describe within-regime transitions, while the paired difference is algebraically a matched comparison of POST similarities.

After freezing the primary Study~2 results, the research program prospectively specified validation on the existing outputs. An independent identity-similarity evaluator and a blinded human panel provide directional concordance on different measurement scales. Models using measured face size, sharpness, detector confidence, and pose provide a separate, supportive sensitivity analysis within the original \buffalo\ outcome. This distinction prevents treating adjustment of one evaluator as another independent measurement of identity. The validation adds evidence about the measured comparison, without constituting replication in a new generator, corpus, or participant population.

The contribution is thus a controlled empirical characterization of person-wise change in one implementation, accompanied by transparent availability accounting and bounded interpretation. We contribute no new generation module, repair policy, or predictor. The following sections position this comparison against existing preservation work, describe the frozen design and estimators, present the sequential empirical findings, and distinguish independent validation from supportive sensitivity. The Supplement provides detailed model specifications, numerical tables, additional statistical figures, and ethics and data-availability disclosures.

%% file: sections/related_work.tex
\section{Related Work}
\label{sec:related}
\paragraph{Identity personalization and multi-subject generation.}
Textual Inversion and DreamBooth connect subject examples to a text-to-image generator through learned embeddings or adaptation \citep{gal2022image,ruiz2023dreambooth}. Encoder-based methods such as IP-Adapter, PhotoMaker, InstantID, and PuLID provide reference-driven conditioning without per-subject optimization in the evaluated generation process \citep{ye2023ipadapter,li2023photomaker,wang2024instantid,guo2024pulid}. These works establish the wider personalization setting. Our study uses an existing generator and its existing identity representations. It does not compare personalization algorithms or propose a replacement for their conditioning mechanisms, and the ability to render a reference identity is distinct from the stability of another person's measured transition.

Multi-subject methods address interference within shared synthesis. Custom Diffusion supports multi-concept customization; FastComposer uses localized attention for multi-subject generation; Cones~2 and Mix-of-Show support multiple subjects or concepts; Subject-Diffusion generates personalized subjects without test-time fine-tuning; and MuDI addresses identity decoupling \citep{kumari2023multi,xiao2023fastcomposer,liu2023cones,gu2023mixofshow,ma2023subjectdiffusion,jang2024mudi}. UniPortrait unifies single- and multi-human personalization through native identity conditioning and routing \citep{ma2024uniportrait}. These precedents make identity separation an established problem. The present evidence concerns the behavior of its single-ID and dual-ID configurations under one matched source-coupled protocol, rather than a claim that explicit multi-identity conditioning is new.

The comparison also differs from an algorithm benchmark. A study of competing architectures can compare each method's preferred configuration, while this design holds the generator implementation fixed and examines two of its existing operational modes. That choice permits a focused matched-output characterization but narrows generalization. Existing work supplies the conditioning machinery and the preservation objective; the present contribution supplies measured behavior, availability accounting, and interpretation of this particular comparison. No superiority claim against the cited multi-subject methods follows from it.

\paragraph{Source preservation and identity restoration.}
DDIM provides deterministic sampling and the inversion foundation used by the frozen implementation \citep{song2020denoising}. Source-preserving editors use attention or feature control, including Prompt-to-Prompt, Plug-and-Play Diffusion Features, and MasaCtrl \citep{hertz2022prompt,tumanyan2022pnp,cao2023masactrl}. GroupDiff addresses group portrait editing, while EditedID addresses facial restoration after editing, including multi-person settings \citep{groupdiff2024,editedid2024}. These works already make preservation and restoration explicit objectives. Here source coupling means a PRE-derived trajectory supplies attention context to a separately initialized output; it does not imply localized inpainting or copy-back of the protected face. A measured protected-person change is therefore an empirical property of the evaluated operation, not evidence that all preservation methods share it.

\paragraph{Multi-human evaluation and retention.}
MultiHuman-Testbench provides dedicated evaluation of multi-human generation \citep{mhtb2025}; MultiBind examines attribute misbinding across subjects \citep{multibind2024}; MPIE-Bench concerns anatomically plausible multi-person interaction editing \citep{lin2026mpiebench}. Evaluation of unchanged content also appears in EdiVal-Agent, while agentic generation and editing frameworks consider verification and retention across operations \citep{tianyu2025edival,ren2026scope,liu2026generationnavigator}. This study specializes evaluation to two prospectively assigned person roles and continuous reference-based facial similarity. It does not measure anatomical plausibility, instruction success, prompt fidelity, or the performance of an iterative repair agent. Person-wise tracking may reveal changes that a target-only evaluation would overlook, but no unmeasured aggregate metric is shown here to fail.

\paragraph{Identity measurement and validation.}
ArcFace supplies the face-recognition basis for the fixed \buffalo\ representation used as the primary evaluator \citep{deng2019arcface,insightface_models}. ID-Sim provides an identity-focused similarity metric on a distinct learned representation \citep{chae2026idsim}. Agreement between evaluators is useful because reference similarity depends on representation and image formation, but their effect magnitudes are not interchangeable. Human comparison provides another measurement perspective, with uncertainty bounded by the sampled cases and fixed panel. SCRFD detection, DIoU spatial association, and 6DRepNet pose estimation serve specific technical roles \citep{guo2021sample,zheng2020distance,hempel2022sixdrepnet}. Detection and correspondence establish which crop is measured; they do not independently verify that the tracked person retains the same identity. Our empirical positioning rests on this limited combination of regime comparison, person-wise transitions, and complementary measurements, without asserting priority for multi-ID preservation itself.

%% file: sections/design_measurement.tex
\section{Design and Measurement}
\label{sec:design}
\subsection{Cohort, roles, and chronology}
The canonical universe contains 375 two-person test cases combining prompt templates and identity assignments, rather than 375 independently sampled prompts. Frozen filters excluded 216 cases, including pilot/development material and nonconforming cases, leaving 159 candidates in a fixed permutation. Sequential accrual used baseline seed 101 and an outcome-independent target-role rule. Candidates 1--129 yielded 101 accepted PRE scenes; reserve ranks 130--159 yielded 18, exhausting the pool at 119 instead of the planned 120. No rescue seed, replacement generation, or reopened exclusion was used. Alternating target roles produced 60 cases targeting person~1 and 59 targeting person~2. Acceptance and correspondence were fixed before identity scoring.

Study~1 generated A1 and A0 realizations for each accepted case at seeds 1101, 2202, and 3303, giving 357 planned outputs per action. Study~2 retained the frozen A1 outputs and prospectively evaluated A2 on the same grid. Its primary results were frozen before the validation program was designed; validation specifications were fixed before their execution on existing outputs. The validation specifications were prospective with respect to validation execution, but followed the primary Study~2 result freeze.

The design unit is a case with its prompt and reference assignment, and the repeated realization unit adds an output seed. Both person roles are measured in each available realization, against their respective invariant references. A matched Study~2 row therefore connects the same case, PRE scene, reference pair, role assignment, and output seed across A1/A2. Reusing A1 does not create an additional independent sample. The two studies answer sequential questions on the accepted cohort: characterization of target-only behavior first, then a prospective comparison with another native regime. The validation subsequently reuses those outputs under additional measurements and specified sensitivity models.

\subsection{Operational regimes and the two-slot mechanism}
Under A1, the accepted PRE scene is inverted with deterministic DDIM \citep{song2020denoising}. An inversion callback maintains the resulting source trajectory in slot~0. The measured output in slot~1 starts from a seeded Gaussian latent and receives source-slot key/value information through UniPortrait's shared self-attention. Whole-canvas denoising proceeds under the original prompt and the target reference alone. It uses 25 DDIM steps, guidance 7.5, face-ID scale 0.7, face-structure scale 0.3, and shared-attention score shift $-0.5$. The measured output is not initialized directly from the PRE inversion latent. Figure~\ref{fig:protocol} separates generation from subsequent correspondence and scoring.

A2 uses the same PRE-derived source trajectory, matched slot~1 seed, prompt, and sampling settings, while supplying both reference portraits in natural prompt order. It activates UniPortrait's native dual-ID configuration. A1 uses one identity, single-ID LoRA weight 1, multi-ID LoRA weight 0, and 16 identity tokens; A2 uses two identities, weights 0 and 1 respectively, and 32 tokens. Reference inclusion, routing, adapters, and token count co-vary. The contrast is therefore between intact native regimes, without component-level causal attribution. Neither action uses spatial identity masks, box conditioning, crop/paste, copy-back, localized inpainting, or an active ControlNet contribution.

A0 is global regeneration from fresh Gaussian noise with both identity references and the original prompt, matched seeds, and the same denoising length and guidance/scales. It omits PRE-derived shared-attention coupling. Its purpose is contextual characterization of the A1 transition relative to an uncoupled generative comparator. It differs from A1 in conditioning and source coupling as well as in the correspondence procedure. The design does not turn A0 into an otherwise identical control for one isolated module, nor does it establish successful target correction under any action.

\input{figures/protocol}

\subsection{Correspondence, continuous identity, and estimands}
SCRFD detection uses the frozen confidence threshold of 0.50 \citep{guo2021sample}. Structural evaluability requires exactly two detected faces; PRE acceptance and A0's technical presentation gate implement this requirement before human correspondence. Three independent blinded raters assign complete reference-to-face bijections for PRE and A0, accepted with at least two-of-three consensus. A1/A2 use the frozen DIoU endpoint to associate POST detections with PRE spatial anchors \citep{zheng2020distance}. Insufficient detections and collisions remain structurally missing. DIoU tracks spatial association, not independently established identity continuity; biometric scores are prohibited from correspondence assignment. Frozen structural rules and their different action-specific availability mechanisms are retained, rather than retrospectively harmonized.

After correspondence, face crops are aligned with the frozen detector keypoints using \path{face_align.norm_crop}. The 512-dimensional \buffalo\ \texttt{w600k\_r50} image and reference embeddings are L2-normalized \citep{deng2019arcface,insightface_models}. Their dot product is cosine similarity. For each role $p\in\{j,k\}$, the reference is fixed across conditions:
\begin{equation}
\ID_p(x)=\langle e_p(x),e_p^{\mathrm{ref}}\rangle,\qquad
\dID_{p,a}=\ID_{p,a}^{\POST}-\ID_p^{\PRE}.
\label{eq:transition}
\end{equation}
Negative change means lower reference similarity. Roles do not encode PASS/FAIL states, and zero is a sign boundary rather than a calibrated preservation threshold. The protected endpoint is primary and the target endpoint secondary in both studies.

In Study~2, the paired role-specific difference for a jointly measurable case and seed is
\begin{align}
D_p&=\dID_{p,\Atwo}-\dID_{p,\Aone} \nonumber\\
&=(\ID_{p,\Atwo}^{\POST}-\ID_p^{\PRE})
 -(\ID_{p,\Aone}^{\POST}-\ID_p^{\PRE}) \nonumber\\
&=\ID_{p,\Atwo}^{\POST}-\ID_{p,\Aone}^{\POST}.
\label{eq:cancellation}
\end{align}
The shared PRE score cancels exactly. PRE remains essential for reporting absolute within-regime transitions and supplies the common source context. It does not add an independent longitudinal estimand to the paired contrast, which is the matched POST difference in reference similarity. Accordingly, a positive $D_p$ establishes a relative comparison; it need not mean A2 improved similarity over PRE.

\subsection{Inference and availability}
Study~1 fits a REML mixed model with six Action$\times$Seed fixed-effect cells, an unstructured two-by-two covariance for action-specific case effects, and action-specific residual variances. Marginal means average the three fitted seed cells equally; their A1$-$A0 difference is the process contrast. Intervals use 5,000 case-cluster bootstrap resamples with NumPy \texttt{RandomState} seed 2029. These fitted means differ from raw descriptive or case-averaged quantities. The Supplement gives the complete specification and frozen variance components.

Study~2 fits $D_p\sim\mathrm{Seed}+(1\mid\mathrm{Case})$ by REML, with categorical seed effects, a random case intercept, and equal seed weighting. Percentile intervals use 5,000 case-cluster resamples with seed 20260921. Its denominator is jointly measurable pairs. The supportive all-available model adds action-specific case effects, a Case:Seed intercept, and action-specific residual variances, using profile REML and Wald intervals. It is an observed-data sensitivity check; it neither imputes unavailable outcomes nor removes measurability conditioning. Complete-replicate and seed-stratified checks retain their original estimators and denominators.

The cluster unit for uncertainty is Case, so repeated seeds are not treated as independent case draws. Equal seed weighting defines the fitted marginal contrast even when availability differs among seeds. Raw summaries may instead weight the observed rows implicitly and are labeled separately. This distinction is necessary when presenting fitted seed means beside seed-specific sample means, and when comparing the primary paired estimate with the all-available specification. These labels and interval types distinguish the reported contrasts.

Table~\ref{tab:design} distinguishes planned outputs, measurable outputs, and paired availability. Missing results remain null rather than negative identity scores. Case clustering accounts for repeated seeds, but intervals remain conditional on the evaluated corpus, accepted PRE scenes, identity assignments, and measurement rules. They do not quantify uncertainty for unseen prompts, identities, or alternative PRE seeds. This scope also applies to the validation analyses described below.
\input{tables/design}

%% file: figures/protocol.tex
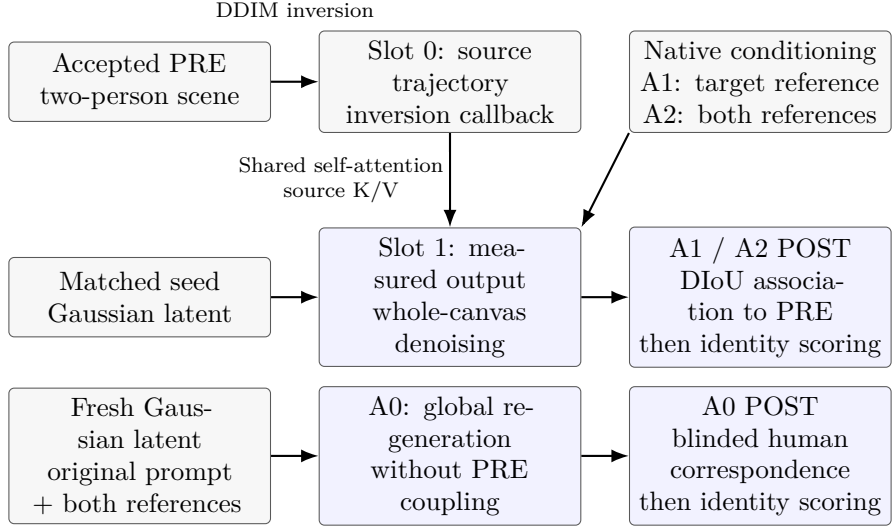
\begin{figure}[t]
\centering
\begin{tikzpicture}[x=1cm,y=1cm,>=Latex,
 box/.style={draw=black!65,rounded corners=2pt,fill=black!3,align=center,font=\small,text width=3.2cm,minimum height=1.05cm},
 output/.style={box,fill=blue!5},arrow/.style={->,thick}]
\node[box] (pre) at (0,3.1) {Accepted PRE\\two-person scene};
\node[box] (source) at (4.1,3.1) {Slot 0: source trajectory\\inversion callback};
\node[box] (config) at (8.2,3.1) {Native conditioning\\A1: target reference\\A2: both references};
\node[box] (noise) at (0,0.25) {Matched seed\\Gaussian latent};
\node[output] (slot) at (4.1,0.25) {Slot 1: measured output\\whole-canvas denoising};
\node[output] (post) at (8.2,0.25) {A1 / A2 POST\\DIoU association to PRE\\then identity scoring};
\draw[arrow] (pre)--(source);
\node[font=\scriptsize] at (2.05,4.05) {DDIM inversion};
\draw[arrow] (noise)--(slot);
\draw[arrow] (source)--node[left,font=\scriptsize,align=center,fill=white,inner sep=2pt]{Shared self-attention\\source K/V}(slot);
\draw[arrow] (config.south west)--(slot.north east);
\draw[arrow] (slot)--(post);
\node[box] (a0in) at (0,-1.85) {Fresh Gaussian latent\\original prompt + both references};
\node[output] (a0) at (4.1,-1.85) {A0: global regeneration\\without PRE coupling};
\node[output] (a0post) at (8.2,-1.85) {A0 POST\\blinded human correspondence\\then identity scoring};
\draw[arrow] (a0in)--(a0);
\draw[arrow] (a0)--(a0post);
\end{tikzpicture}
\caption{Design schematic of the evaluated implementation. A1/A2 supply PRE-derived attention context from source slot~0 to a seeded Gaussian output in slot~1. The reference configuration changes as an intact native regime. A0 has no PRE coupling. DIoU and human correspondence occur after generation and do not condition it. Boxes denote operations rather than identity-preservation guarantees.}
\label{fig:protocol}
\end{figure}

%% file: tables/design.tex
\begin{table}[t]
\centering\small
\caption{Study design and availability. All actions use 119 accepted PRE cases and seeds 1101, 2202, 3303. A1 is reused in Study~2. Joint Study~2 counts partition the 357 planned pairs; a case can contribute several seeds.}
\label{tab:design}
\begin{tabularx}{\linewidth}{Xrrr}
\toprule
Metric & A1 & A0 & A2\\\midrule
Planned outputs per action & 357 & 357 & 357\\
Identity-measurable outputs & 304 & 293 & 319\\
Unavailable outputs & 53 & 64 & 38\\
Correspondence & DIoU & Human consensus & DIoU\\\midrule
\multicolumn{4}{l}{Study~1: 597 measurable action outputs; 257 jointly measurable case--seed pairs.}\\
\multicolumn{4}{l}{Study~2: 287 paired rows from 111 cases; 74 complete three-seed cases.}\\
\multicolumn{4}{l}{Study~2 partition: both 287; A1 only 17; A2 only 32; neither 21.}\\
\multicolumn{4}{l}{Study~2 all-available sensitivity: 623 action observations.}\\
\bottomrule
\end{tabularx}
\end{table}

%% file: sections/empirical_findings.tex
\section{Empirical Findings}
\label{sec:findings}
\subsection{Study 1: broad perturbation under target-only generation}
Study~1 provides the initial characterization of the source-coupled single-ID regime. Among 357 planned A1 outputs, 304 were measurable; A0 contributed 293 of 357, giving 597 action observations. The fitted protected-person mean transition under A1 was $-0.567904$, compared with $-0.010801$ under A0. The A1$-$A0 process contrast was $-0.557103$ with 95\% case-cluster bootstrap CI $[-0.582850,-0.530484]$. The near-zero A0 mean contextualizes the pronounced A1 decrease, while its interval includes zero. Table~\ref{tab:primary} reports all primary and secondary fitted estimates, intervals, and bootstrap standard errors; Figure~\ref{fig:protected} displays the protected-person distributions.

The target also showed a negative A1 mean transition, $-0.236904$, compared with $-0.021632$ under A0. Its A1$-$A0 contrast was $-0.215272$ (95\% CI $[-0.245432,-0.184669]$). Thus the target-directed operation did not establish successful average target repair. Relative to the accepted PRE scenes, both reference similarities decreased in the fitted A1 means. The protected-person result is a substantial continuous decrease under the evaluated operator, rather than a categorical finding that an identity was destroyed. A0 and A1's different correspondence and availability procedures also remain part of the comparison's interpretation.

The joint signs sharpen this interpretation without adding a task-success threshold. All 304 measurable A1 realizations had negative protected transitions. Of these, 275 (90.46\%) also had nonpositive target transitions, and 29 (9.54\%) had positive target transitions. The dominant joint pattern was therefore simultaneous deterioration in reference similarity for both roles, while a minority showed a positive target transition. These observations do not justify presenting the aggregate protected decrease as the collateral price of established successful repair. They also do not imply that every target transition was negative. The full sign partitions and descriptive associations are retained in Supplement Table~S4.

A0 observations were distributed across all four sign quadrants: 45 target-positive/protected-negative, 77 target-positive/protected-nonnegative, 118 target-nonpositive/protected-negative, and 53 target-nonpositive/protected-nonnegative outputs. The descriptive target/protected correlations were close to zero for A1 (Pearson $r=0.090640$, Spearman $\rho=0.026326$) and positive for A0 ($r=0.491008$, $\rho=0.452616$). These coefficients summarize realized variation without cluster-adjusted inference or causal interpretation. They supplement the fitted mean contrasts rather than explain their mechanism. In particular, neither quadrant occupancy nor correlation converts the continuous endpoint into a diagnosis of identity swapping or a calibrated preservation decision.

Frozen sensitivity summaries agree in direction. Restricting to the 110 cases with at least one measurable replicate under each action yielded protected/target contrasts of $-0.561320$/$-0.214189$; the 57 cases with complete three-seed coverage under both actions yielded $-0.586572$/$-0.212455$. Seed-specific protected contrasts were $-0.570254$, $-0.572318$, and $-0.564490$ at seeds 1101, 2202, and 3303. These checks describe different observed subsets and summaries; they do not replace the full REML estimator or demonstrate that unavailable outputs would behave similarly. Supplement Tables~S1--S3 retain coverage, complete intervals, medians, variance components, and frozen convergence diagnostics.

\input{tables/primary}

\subsection{Study 2: an intact native-regime difference}
A2 produced 319 measurable outputs from 357 planned realizations, compared with A1's frozen 304. Exactly 287 case--seed pairs were measurable under both actions, contributed by 111 cases; 74 cases contributed all three pairs. On this paired population, the fitted protected contrast was $+0.512918$ (95\% CI $[0.471434,0.550259]$), and the target contrast was $+0.158853$ ($[0.118903,0.197523]$). Both are equal-seed-weighted REML estimates. They indicate more positive mean transitions under A2, equivalently higher matched POST reference similarity by Equation~\ref{eq:cancellation}. The positive target mean contrast accompanies the positive protected mean contrast; it is not a claim that every realization pair improved for both people.

Both paired models retained non-singular random-case-intercept specifications. Their frozen variances and available convergence diagnostics are reported in Supplement Table~S6.

The direct descriptive comparison on the identical 287-pair denominator shows why relative improvement should be distinguished from absolute preservation. Protected means were $-0.589143$ for A1 and $-0.072655$ for A2; target means were $-0.241357$ and $-0.081344$. The corresponding unweighted paired differences, $+0.516488$ and $+0.160013$, differ from the fitted primary contrasts and are not substituted for them. A2's mean absolute transitions remained negative on this paired sample, even though its protected median was $+0.000230$. Supplement Table~S9 preserves all distribution summaries with separate all-A2, paired, and all-A1 denominators.

Complete three-seed paired cases yielded protected/target contrasts of $+0.526683$/$+0.169166$ across 74 cases and 222 pairs. The seed-stratified raw paired means were $+0.522529$/$+0.185371$ for seed 1101, $+0.520796$/$+0.135895$ for 2202, and $+0.506219$/$+0.159552$ for 3303. All intervals remained above zero. These raw stratum summaries are distinct from the fitted seed marginal means in Supplement Table~S6. The supportive all-available model used 623 action observations and yielded contrasts of $+0.498569$/$+0.157553$ with Wald, rather than bootstrap, intervals. Its A2 fitted mean transitions remained $-0.070088$/$-0.079556$. These observed-data checks support directional consistency without recovering unobserved outcomes.

Across all 319 measurable A2 outputs, the protected and target raw mean transitions were $-0.070549$ and $-0.080126$. Their joint sign quadrants contained 46 target-positive/protected-negative, 91 both-nonnegative with target-positive, 114 target-nonpositive/protected-negative, and 68 target-nonpositive/protected-nonnegative realizations. These are sign partitions of continuous measurements, not categorical success/failure labels. The strong descriptive Pearson association under A2 ($r=0.826727$) also does not establish shared causation or supply cluster-adjusted correlation inference. The remaining distribution plots and all exploratory coefficients are in the Supplement.

The all-A2 protected median was $-0.000015$, with interquartile bounds $[-0.064465,+0.035160]$; the target median was $-0.015024$, with bounds $[-0.077919,+0.029863]$. Their standard deviations were $0.208261$ and $0.213588$. A protected distribution centered near zero can coexist with a negative mean and a positive relative contrast against A1. The distinct 319-output and 287-pair populations also explain why their summaries should not be merged into one absolute-change claim. The full ranges and paired distributions in Supplement Table~S9 allow inspection of this variation while retaining the frozen denominator for each field. Figure~\ref{fig:qualitative} illustrates these regime dynamics qualitatively across two contrastive cases: Case~054 demonstrates relative attenuation of protected identity degradation under A2 ($D_j = +0.605, D_k = +0.415$), whereas Case~051 illustrates an atypical negative contrast ($D_j = -0.128, D_k = -0.201$) accompanied by visible disagreement between facial appearance and the frozen spatial assignment slots. As emphasized by the metrics, these continuous deltas reflect relative changes rather than categorical preservation.

\input{figures/protected_results}
\input{figures/qualitative}

\subsection{Availability bounds the findings}
Within Study~2's 357 planned pairs, 17 were measurable only under A1, 32 only under A2, and 21 under neither. A1's 53 missing outputs comprised 35 insufficient-face detections and 18 tracking collisions; A2's 38 comprised 9 insufficient-face detections and 29 collisions. In Study~1, A0 additionally had 49 technical gate failures and 15 unresolved human correspondences. These mechanisms differ by action. The primary paired inference describes jointly measurable realizations, rather than all planned generations, and no missing observation was converted into an identity penalty. Complete-case and all-available sensitivity findings do not establish unbiasedness or remove this selection boundary. The observed regime difference remains a comparison of the native configurations in this measurable cohort.

%% file: tables/primary.tex
\begin{table}[t]
\centering\small
\setlength{\tabcolsep}{4pt}
\caption{Primary and secondary fitted \buffalo\ estimates. Study~1 uses 597 measurable action observations in the 119-case design (A1 304, A0 293); Study~2 uses 287 jointly measurable pairs from 111 cases. All rows use equal seed weights and 5,000 case-cluster bootstrap resamples. Full variance and sensitivity records are in Supplement Tables~S2, S3, S6--S8.}
\label{tab:primary}
\begin{tabularx}{\linewidth}{lXrrr}
\toprule
Study / role & Estimand & Estimate & 95\% bootstrap CI & Bootstrap SE\\\midrule
1 / Protected $j$ & A1 transition & $-0.567904$ & $[-0.597297,-0.536832]$ & $0.015604$\\
& A0 transition & $-0.010801$ & $[-0.030899,+0.011336]$ & $0.010925$\\
& A1$-$A0 & $-0.557103$ & $[-0.582850,-0.530484]$ & $0.013219$\\\addlinespace
1 / Target $k$ & A1 transition & $-0.236904$ & $[-0.265309,-0.209629]$ & $0.014251$\\
& A0 transition & $-0.021632$ & $[-0.038848,-0.003419]$ & $0.009013$\\
& A1$-$A0 & $-0.215272$ & $[-0.245432,-0.184669]$ & $0.015603$\\\addlinespace
2 / Protected $j$ & Paired A2$-$A1 & $+0.512918$ & $[+0.471434,+0.550259]$ & $0.020057$\\
2 / Target $k$ & Paired A2$-$A1 & $+0.158853$ & $[+0.118903,+0.197523]$ & $0.020346$\\
\bottomrule
\end{tabularx}
\end{table}

%% file: figures/protected_results.tex
\begin{figure}[t]
\centering
\begin{subfigure}{0.70\linewidth}
\centering\includegraphics[width=\linewidth]{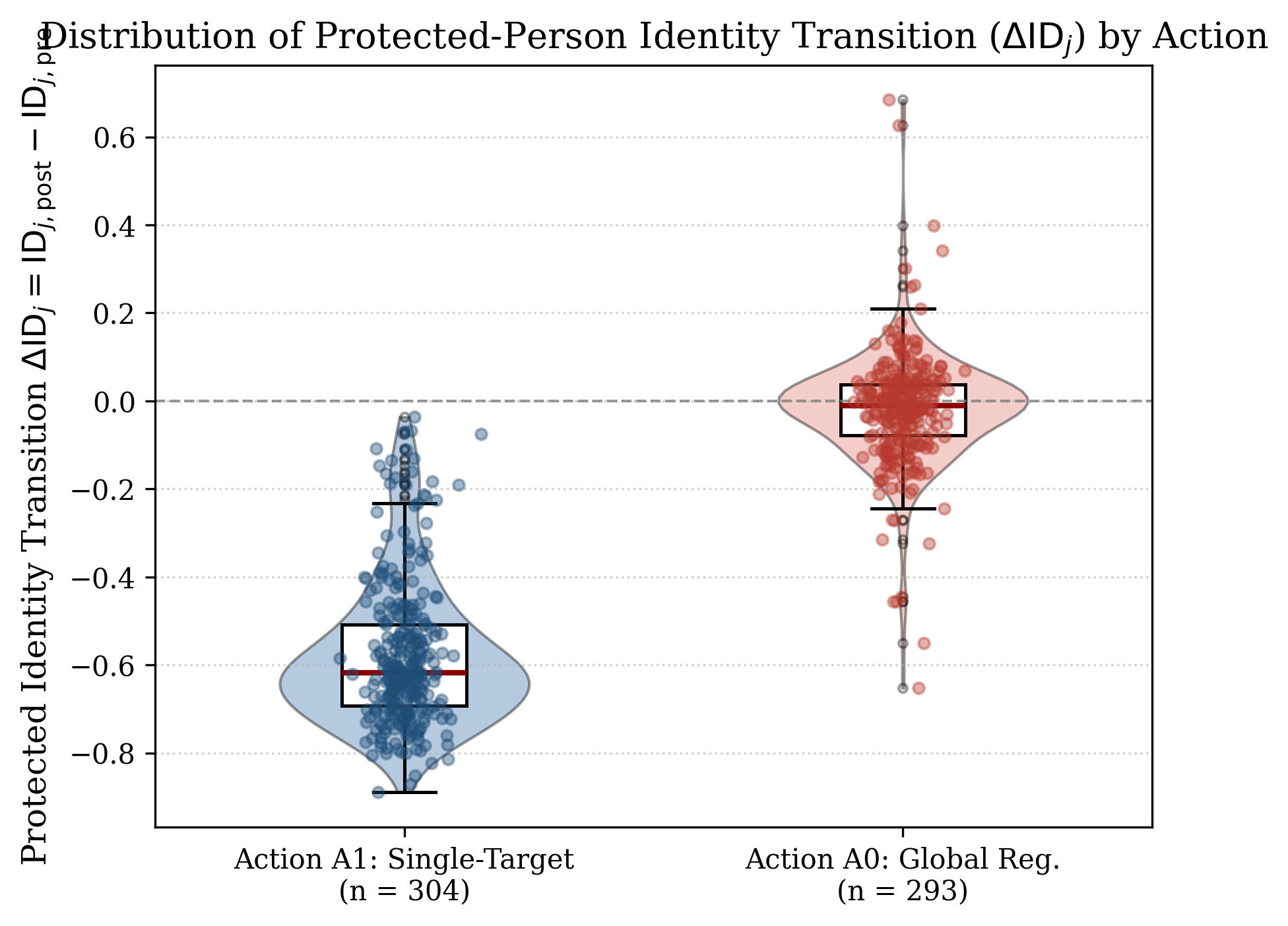}
\caption{Study~1 protected transitions: A1 versus A0.}
\end{subfigure}
\par\vspace{8pt}
\begin{subfigure}{0.98\linewidth}
\centering\includegraphics[width=\linewidth]{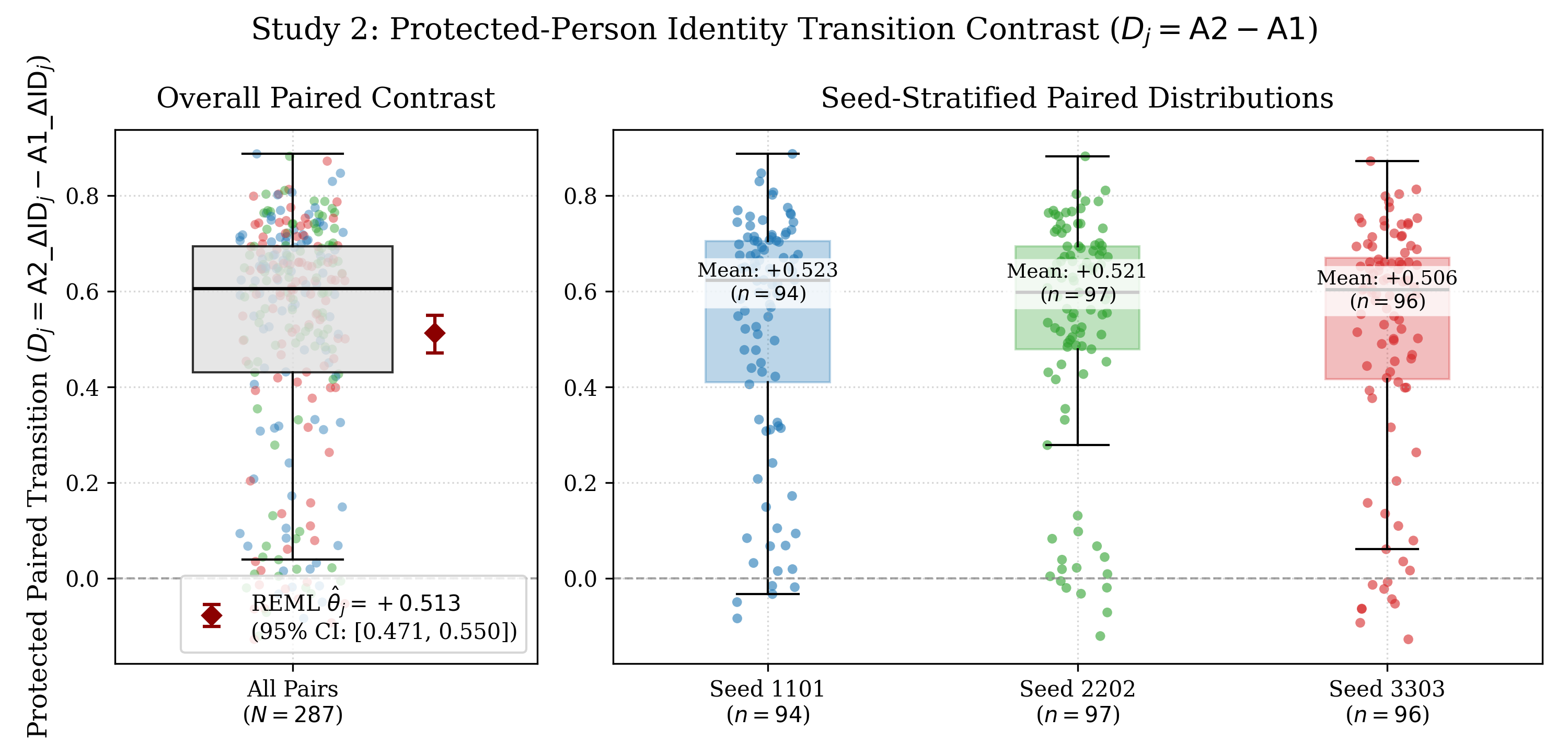}
\caption{Study~2 protected paired contrast: A2$-$A1.}
\end{subfigure}
\caption{Protected-person evidence across the sequential studies. (a) Distributions of $\dID_j$ for A1 ($n=304$) and A0 ($n=293$). (b) 287 jointly measurable paired contrasts, their fitted estimate and 95\% case-cluster bootstrap interval, and seed strata ($n=94,97,96$). The panels have different endpoints and scales: within-action change in (a), paired between-action difference in (b).}
\label{fig:protected}
\end{figure}

%% file: figures/qualitative.tex
\begin{figure}[t]
\centering\includegraphics[width=\linewidth]{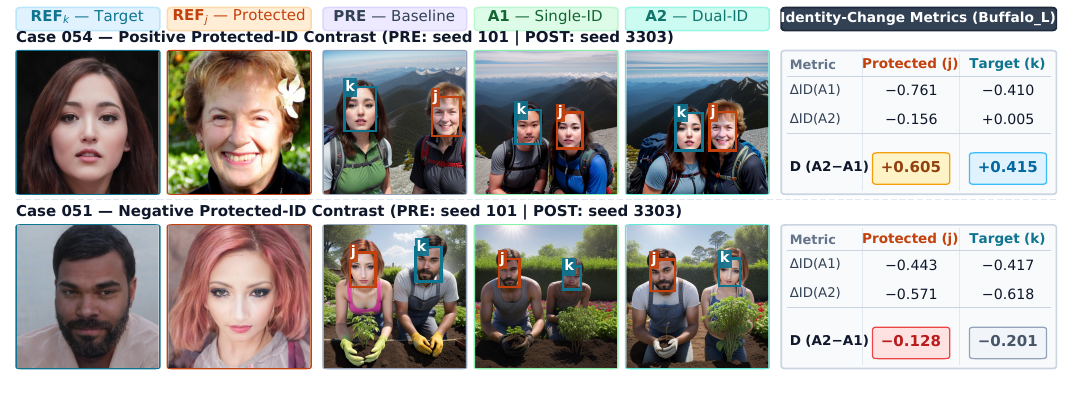}
\caption{\textbf{Qualitative comparison of person-wise identity stability under single-identity (A1; target-only) and dual-identity (A2; target and protected) conditioning regimes across two illustrative cases sharing matched sampling seeds (POST seed 3303; baseline PRE seed 101).}
Columns 1--2 display target ($\mathbf{I}_k^{\mathrm{ref}}$) and protected ($\mathbf{I}_j^{\mathrm{ref}}$) reference portraits.
Column 3 shows the unconditioned baseline scene ($\mathbf{PRE}$).
Columns 4--5 show realizations under A1 and A2.
Column 6 reports \buffalo\ identity-similarity changes: transition deltas $\dID(A) = \ID_{\mathrm{post}}(A) - \ID_{\mathrm{pre}}$ and paired contrasts $D = \dID(\mathrm{A2}) - \dID(\mathrm{A1})$.
Colored boxes and corner badges ($k$: teal; $j$: coral) indicate evaluated spatial assignment slots, not independently established visual identity matches.
In Case 054, omitting the protected subject under A1 induces substantial identity degradation ($\dID_j = -0.761$), whereas A2 attenuates this loss ($D_j = +0.605, D_k = +0.415$).
In Case 051, dual conditioning produces negative contrast ($D_j = -0.128, D_k = -0.201$), accompanied by visible disagreement between facial appearance and the frozen spatial assignment slots.
These qualitative examples illustrate regime dynamics and do not establish categorical identity preservation or population-level repair success.}
\label{fig:qualitative}
\end{figure}

%% file: sections/validation.tex
\section{Independent and Supportive Validation}
\label{sec:validation}
Validation was designed after the primary Study~2 result freeze and before validation execution on the frozen outputs. It asks whether the observed direction agrees with another identity representation and human judgment, and separately whether the original outcome is sensitive to specified measured nuisances. These roles are not interchangeable. Figure~\ref{fig:validation} presents the validation results on separate scales; Supplement Table~S10 reports all estimates and intervals. None of these analyses adds a new generator or independent study cohort.

\subsection{V1: independent metric concordance}
ID-Sim uses the frozen DINOv3 ViT-L/16 \texttt{cls\_patch} checkpoint and \texttt{cls} mode \citep{chae2026idsim}. Its distance is oriented as PRE distance minus POST distance, so a positive transition denotes increased reference consistency. On the same 287 pairs from 111 cases, its protected A2$-$A1 contrast was $+0.231351$ (95\% case-cluster CI $[0.207440,0.254732]$), and the target contrast was $+0.072272$ ($[0.048170,0.097237]$). The frozen paired model and 5,000-resample case bootstrap used seed 20260927. Both directions agree with \buffalo. This is independent-evaluator concordance on shared outputs, not replication of the experiment. Distinct distance and cosine scales preclude direct effect-size comparison.

\subsection{V2: blinded human identity comparison}
The prespecified human-validation protocol selected 80 of the 111 contributing cases, balanced 40/40 across target-role assignments. One jointly measurable seed per case was chosen by the smallest frozen hash, and both person roles were presented. Five final primary raters independently judged 160 packets, giving 800 responses; candidate orientation was balanced. The final panel was administratively amended without outcome access and locked before condition decoding. Blinding concealed regimes, case identifiers, and biometric scores. For an evaluable packet with at least three judgeable responses, $\mathcal H=(N_{\mathrm{A2}}-N_{\mathrm{A1}})/N_{\mathrm{judgeable}}$; ``no clear difference'' contributes zero and remains judgeable.

All 80 protected packets were evaluable. Of 400 responses, 370 favored A2, none A1, four reported no clear difference, and 26 were unable to judge. Mean $\mathcal H$ was $+0.986667$ (95\% CI $[0.965000,1.000000]$). For the target, 77 packets were evaluable; the 400 responses comprised 184 A2, 34 A1, 151 no clear difference, and 31 unable judgments. Mean $\mathcal H$ was $+0.401948$ ($[0.304816,0.498673]$). The larger target tie fraction, 37.75\% versus 1.0\%, tempers an undifferentiated claim of equally strong perceptual evidence for both roles.

The 5,000-resample case bootstrap used seed 20260928 and is conditional on the fixed panel. Its mean pairwise agreement was 67.75\%, with Fleiss' $\kappa=0.5308$. Leave-one-rater-out means remained positive: protected $+0.9844$ to $+0.9906$, target $+0.3258$ to $+0.4470$. These checks describe panel-member influence, without inference to a population of raters. Full response accounting and the documented instruction scope are retained in Supplement Section~S4.

\subsection{V3: supportive within-evaluator sensitivity}
V3 retains the paired \buffalo\ outcome and adjusts role-specific models for standardized A2$-$A1 differences in log face area, log-transformed Laplacian sharpness, and detector confidence. Pose-augmented models add absolute yaw and pitch from frozen 6DRepNet outputs \citep{hempel2022sixdrepnet}. Core protected/target contrasts were $+0.511637$/$+0.158524$; pose-augmented contrasts were $+0.511421$/$+0.158543$, with all reported intervals above zero. Evaluation at standardized covariates zero denotes the mean observed nuisance profile, not physical equality between actions. These models show directional stability under specified measured covariates within the same evaluator. They neither establish independent identity validity nor eliminate quality effects, confounding, or unmeasured alternative explanations.

The core protected interval was $[0.471271,0.548896]$ and target interval $[0.122081,0.192932]$; pose-augmented intervals were $[0.471351,0.550273]$ and $[0.123847,0.192405]$, respectively. Profile REML and a random case intercept retained the frozen estimation structure. Percentile intervals used 5,000 case-cluster resamples with NumPy \texttt{RandomState} seed 20260929, keeping repeated sampled case occurrences as independent bootstrap blocks. Supplement Section~S5 provides the complete model specification and higher-precision estimates. The supportive interpretation follows from the measured specification, rather than asserting that image-formation differences have been physically equalized.
\input{figures/validation}

%% file: figures/validation.tex
\begin{figure}[t]
\centering\includegraphics[width=0.97\linewidth]{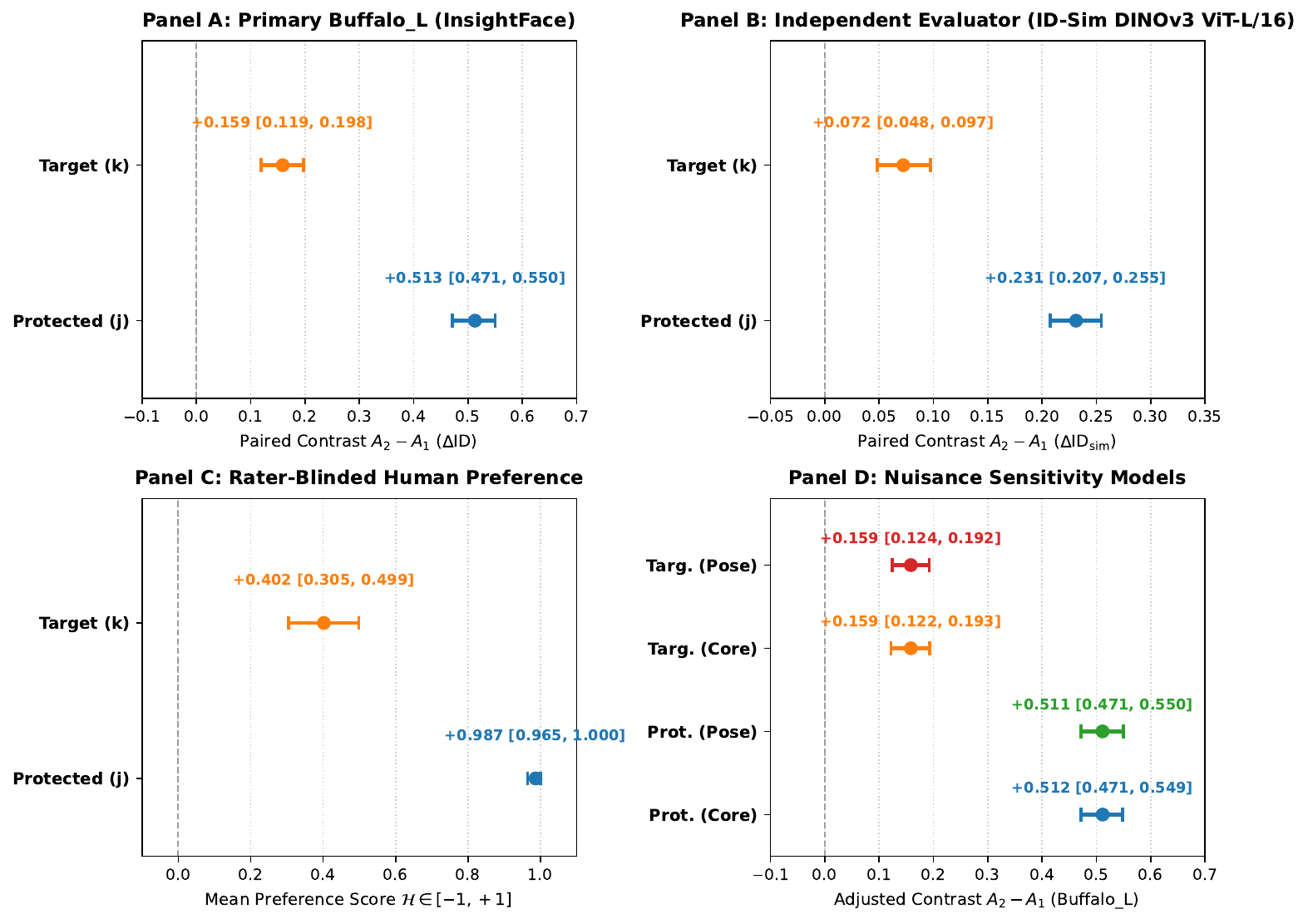}
\caption{Validation on the same generated outputs. Panel~A is the primary \buffalo\ paired contrast; B is independent metric concordance (ID-Sim); C is blinded human preference ($\mathcal H$); D is \textbf{supportive sensitivity within \buffalo}, not an independent evaluator. Error bars are 95\% case-cluster bootstrap intervals. Each panel has its own horizontal scale; magnitudes across metrics and human preference must not be compared. V1/V2 provide independent measurement perspectives on the same outputs, whereas V3 retains the primary outcome and varies its measured-covariate specification.}
\label{fig:validation}
\end{figure}

%% file: sections/discussion.tex
\section{Discussion / Limitations}
\label{sec:discussion}
The sequential findings distinguish three statements that are easily conflated. First, target-only source-coupled generation in the evaluated implementation produced negative mean similarity transitions for both roles. Second, an intact dual-ID regime produced more positive protected and target contrasts on jointly measurable outputs. Third, independent metric and human measurements agreed with the direction of that relative comparison. Together they support bounded regime dependence in measured person-wise identity behavior. They do not establish successful repair, complete identity invariance, or a new preservation mechanism. The descriptive and supportive A2 means remaining below PRE further constrain a categorical preservation reading.

The shared-baseline algebra also determines the contribution's scope. Within-action transitions explain how an output differs from the accepted PRE scene, providing the context necessary to distinguish attenuation from recovery. The paired A2$-$A1 difference itself reduces to a POST similarity comparison against the same reference. Its scientific value lies in the matched source/seed design, prospective role assignment, observed effect, and complementary validation, rather than a new temporal contrast that is statistically independent of matched POST comparison. Reporting the cancellation alongside the transition definition makes this relation explicit and prevents a formal reparameterization from being presented as algorithmic or estimand novelty.

Mechanistic attribution remains unresolved. Supplying the protected reference in A2 co-occurs with native routing, adapter weights, and additional identity tokens. These intact configurations are meaningful operational choices, but the data cannot assign the contrast to any one component. Source coupling also does not imply that spatially tracked identity is causally preserved. DIoU follows PRE-associated boxes, and the embeddings measure reference similarity at those assigned crops. Identity swap and identity drift were not independently adjudicated as separate phenomena. No conclusion about their relative prevalence or mechanism follows from spatial correspondence alone.

Availability is another substantive boundary. Acceptance selected structurally evaluable PRE scenes at one baseline seed. Subsequent missingness arose through action-specific detection, tracking, and human-correspondence mechanisms. The paired analysis conditions on joint measurability; sensitivity analyses condition on other observed subsets or on all available observations. Their agreement does not reveal the outcomes of missing generations or prove unbiased selection. Case bootstrap intervals account for repeated realizations within the evaluated cases. They do not turn those cases into independently sampled prompt templates or provide inference over the wider identity, prompt, or rejected-PRE populations.

Generalization is limited to UniPortrait with the evaluated Stable Diffusion~1.5 implementation, two-person scenes, and the fixed output seeds. Other generators and three-or-more-person compositions were not tested. The operation is whole-canvas source-coupled generation, rather than demonstrated successful localized editing. No instruction-success endpoint, automated selection rule, risk predictor, or component ablation was evaluated. Thus a practical implication is to monitor each relevant person when assessing a comparable operation; it is not a validated deployment policy. Target-focused or whole-image evaluation may miss person-specific change, but this study did not measure their relative diagnostic accuracy.

Validation narrows some measurement concerns while retaining its own limits. ID-Sim and \buffalo\ are different learned representations operating on the same frozen outputs; agreement does not make either a calibrated identity probability. Human evidence is based on 80 selected cases and five fixed raters, with fewer evaluable target packets and substantially more target ties. Its intervals and omission checks are conditional on that panel. V3 retains \buffalo\ and assesses specified measured nuisances at their observed mean profile. It provides sensitivity evidence only; unmeasured image effects and evaluator-specific behavior remain possible. The concordance classification should therefore be read as a directional protocol summary, not three independent replications.

Ethics and release constraints also remain explicit. Reference portraits come from the benchmark's documented FFHQ, CelebA-HQ, and SFHQ sources; PRE and POST scenes are synthetic. No rater facial images or biometric measurements were collected, and scientific responses use opaque identifiers. Institutional ethics/consent documentation is pending determination and must be finalized before submission; no approval or exemption is asserted. A curated scientific reproducibility release remains pending and is subject to material-specific licensing, redistribution permissions, and rater privacy.

%% file: sections/conclusion.tex
\section{Conclusion}
\label{sec:conclusion}
In the evaluated two-person UniPortrait setting, source-coupled single-ID target-only generation produced broad negative mean identity-similarity transitions, while native dual-ID generation yielded more positive protected and target contrasts among jointly measurable pairs, with the shared PRE term canceling in that matched comparison. Independent metric and blinded human judgments agreed directionally on the same frozen outputs, and within-\buffalo\ nuisance models supplied supportive sensitivity evidence, bounded by the accepted cohort, fixed panel, measurement rules, and observed-data availability. This empirical characterization motivates person-wise evaluation of co-present individuals in comparable operations, without establishing successful localized repair, categorical identity preservation, component causality, generalization to other generators or larger groups, or a new generation algorithm.

%% file: supplement/methods.tex
\section{Detailed design, implementation, and correspondence}
\label{sec:s_design}
\subsection{Corpus and sequential accrual}
The canonical MultiHuman-Testbench universe contains 375 two-person test cases \citep{mhtb2025}. A case combines a prompt template with assigned identity references; templates can recur across cases. The frozen exclusion registry sequestered 72 pilot/development cases and excluded nonconforming prompts, yielding 216 excluded cases overall and 159 eligible candidates. The eligible pool was permuted with fixed PRNG seed 2028. PRE images used baseline seed 101. The original target was 120 accepted PRE cases, but prospective exhaustion of the 159-candidate pool produced 119. Candidates 1--129 contributed 101 accepted cases and reserve ranks 130--159 contributed 18; no candidate remained unused. The shortfall did not result from post-hoc sample pruning. Rescue seeds, replacement generations, ad-hoc pool extensions, and reopened pilot exclusions were prohibited.

The alternating rank rule assigned 60 cases to Target Person~1 and 59 to Target Person~2. Roles were fixed independently of PRE biometric fidelity. No target-failure or protected-pass threshold was imposed. The accepted PRE cohort is the baseline population for both studies; rejected PRE candidates and other baseline seeds are outside the study's inferential scope. Study~1's grid was $119\times2\times3=714$ planned outputs, 357 each for A1/A0 at seeds 1101, 2202, 3303. Study~2 added 357 prospective A2 outputs at the same case--seed grid and reused the frozen A1 outputs.

\subsection{Exact operational settings}
A1 is UniPortrait's native single-ID target-only regime \citep{ma2024uniportrait}. Deterministic DDIM inversion of PRE supplies a source trajectory maintained in slot~0 by the inversion callback \citep{song2020denoising}. Slot~1 begins with its output seed's Gaussian latent. Shared self-attention conveys source-slot key/value information to the measured slot~1 output during global, unmasked denoising. Slot~1 is not initialized from the PRE inversion latent. Both source-coupled actions use 25 DDIM steps, guidance scale 7.5, face-ID scale 0.7, face-structure scale 0.3, enabled shared attention, and shared score shift $-0.5$.

In A1, only the target reference is supplied: \path{num_faceids=1}, \path{lora_scale=1.0}, \path{multi_id_lora_scale=0.0}, and 16 identity tokens. In A2, both references are supplied in natural prompt order without target-first reordering: \path{num_faceids=2}, \path{lora_scale=0.0}, \path{multi_id_lora_scale=1.0}, and 32 identity tokens, activating native multi-ID routing. Thus protected-reference availability, routing, adapters, and token count change together. A0 uses fresh Gaussian noise, both references, and the original prompt, with the matched output seed, scales, guidance, and step count, without PRE-derived shared-attention coupling.

No spatial face-ID mask (\texttt{faceid\_mask=None}), IP mask (\texttt{ip\_mask=None}), active ControlNet contribution, crop/paste composition, protected-face copy-back, or localized inpainting is used. Neither person boxes nor DIoU condition generation. Main Figure~1 summarizes these operations.

\subsection{Four-stage correspondence and measurement}
Existence is assessed with SCRFD-10G at detection confidence $\ge0.50$ \citep{guo2021sample}. Structural evaluability requires exactly two detected faces. PRE and A0 are presented to three independent blinded raters for complete bijective reference-to-detection mappings, with at least two-of-three consensus required. For A1/A2, the frozen DIoU endpoint associates PRE spatial anchors with POST detections \citep{zheng2020distance}. Collisions, ambiguities, and track loss remain null; recognition embeddings never break ties or assign correspondence. These action-specific procedures and their original ledger categories are retained. They are not retrospectively replaced by one common correspondence algorithm.

Only after correspondence is cryptographically fixed does \buffalo\ scoring occur \citep{insightface_models,deng2019arcface}. \path{face_align.norm_crop} aligns each recognition crop with frozen detector keypoints. Image and fixed-reference 512-dimensional \texttt{w600k\_r50} embeddings are L2-normalized, making their dot product cosine similarity. Each role's reference is invariant across PRE and actions. Continuous $\dID_p=\ID_p^{\POST}-\ID_p^{\PRE}$ is reported without a categorical preservation threshold. Source-associated spatial tracking does not independently establish identity continuity or distinguish identity drift from a biometric identity swap.

\subsection{Availability mechanisms}
Study~1's A1 had 304 resolved outputs and 53 structural failures: 35 with fewer than two faces and 18 collisions. A0 had 308 exactly-two-face technical gate passes; 293 achieved consensus and 15 remained unresolved. Its 49 technical failures comprised 9 one-face and 40 excess-face detections. Coverage was 116 cases with at least one A1 replicate, 113 with at least one A0 replicate, and 110 with at least one under each; 80/78/57 cases had complete three-seed A1/A0/both coverage, respectively. Exactly 257 of 357 case--seed pairs were measurable under both Study~1 actions.

A2 had 319 resolved outputs and 38 structural failures: 29 collisions and 9 insufficient-face detections. Study~2's joint partition was 287 both-measurable, 17 A1-only, 32 A2-only, and 21 neither, with 111 contributing paired cases and 74 complete three-seed cases. The all-available action denominator was 623. Missing outcomes were not imputed, replaced, or converted to negative identity values. These procedures condition the observed populations; sensitivity agreement does not establish unbiasedness for all planned outputs.

%% file: supplement/models.tex
\section{Complete estimators and sensitivity definitions}
\label{sec:s_models}
\subsection{Study 1: six-cell REML estimator}
For role $p\in\{j,k\}$, action $a\in\{\Azero,\Aone\}$, case $c$, and seed $s\in S=\{1101,2202,3303\}$, the frozen model is
\begin{equation}
\dID_{p,c,a,s}=\mu_{p,a,s}+b_{p,c,a}+\epsilon_{p,c,a,s}.
\end{equation}
There are six Action$\times$Seed fixed-effect cells. The two action-specific case effects have an unstructured covariance matrix,
\begin{equation}
\begin{pmatrix}b_{p,c,\Azero}\\b_{p,c,\Aone}\end{pmatrix}
\sim\mathcal N\left(0,\begin{pmatrix}\tau^2_{p,\Azero}&\tau_{p,\Azero,\Aone}\\\tau_{p,\Azero,\Aone}&\tau^2_{p,\Aone}\end{pmatrix}\right),
\qquad \operatorname{Var}(\epsilon_{p,c,a,s})=\sigma^2_{p,a}.
\end{equation}
REML estimates the fixed cells and variance structure. The reported equal-seed marginals are
\begin{equation}
\hat\theta_{a,p}=\frac13\sum_{s\in S}\hat\mu_{p,a,s},\qquad
\hat\theta_{D,p}=\hat\theta_{\Aone,p}-\hat\theta_{\Azero,p}.
\end{equation}
Uncertainty uses the frozen 5,000-resample case-cluster bootstrap and NumPy \texttt{RandomState(2029)}. These are fitted marginal estimators, distinct from raw case means in descriptive or paired sensitivity summaries. Table~S2 preserves all variance components, covariance, correlation, residual ratios, bootstrap standard errors, and intervals.

S1 uses the 110 cases with at least one evaluable realization under each action. S2 uses the 57 cases with complete three-seed availability under both actions. S3 gives seed-specific paired strata, with 82, 83, and 92 paired cases at seeds 1101, 2202, and 3303. Table~S3 retains the original mean/SE/CI/median fields for both roles. The protected model converged with L-BFGS-B in 14 iterations, the target model in 13; both had 5,000 successful bootstrap fits out of 5,000 attempted and 0.00\% failure, below the frozen 1.0\% tolerance.

\subsection{Study 2: paired model and PRE cancellation}
The paired endpoint uses only jointly measurable observations:
\begin{align}
D_{p,c,s}&=\dID_{p,c,s,\Atwo}-\dID_{p,c,s,\Aone}\nonumber\\
&=(\ID^{\POST}_{p,c,s,\Atwo}-\ID^{\PRE}_{p,c})-(\ID^{\POST}_{p,c,s,\Aone}-\ID^{\PRE}_{p,c})\nonumber\\
&=\ID^{\POST}_{p,c,s,\Atwo}-\ID^{\POST}_{p,c,s,\Aone}.
\end{align}
The common PRE term cancels. Reporting absolute within-regime transitions still requires PRE, but the paired contrast is a matched POST comparison, not an independent new longitudinal estimand. The frozen REML model is
\begin{equation}
D_{p,c,s}=\alpha_{p,s}+u_{p,c}+e_{p,c,s},\qquad
D_p\sim\mathrm{Seed}+(1\mid\mathrm{Case}),\qquad
\hat\theta_{\mathrm{paired},p}=\frac13\sum_{s\in S}\hat\alpha_{p,s}.
\end{equation}
Categorical seed effects receive equal weights. Percentile intervals use 5,000 case-cluster resamples with seed 20260921. Tables~S5--S7 preserve accounting, fitted seed marginals, raw seed summaries, random-intercept/residual variances, and the historical 5,000/5,000 protected bootstrap success record. Primary $j$ and secondary $k$ retain the same 287 pairs from 111 cases.

Study~2 S1 uses all three jointly measurable seeds in 74 cases (222 pairs). S2 uses seed-specific paired strata of 94, 97, and 96 pairs. Their raw means differ from the fitted seed marginals; neither replaces the primary equal-weighted REML estimator. S3 is the supportive all-available specification, with 623 observed action outputs:
\begin{equation}
\dID_p\sim\mathrm{Action}\times\mathrm{Seed}+(1+\mathrm{Action}\mid\mathrm{Case})+(1\mid\mathrm{Case:Seed}).
\end{equation}
Profile REML uses an unstructured two-by-two action-specific case covariance, a separate Case:Seed random intercept, and action-specific residual variances. Table~S8 retains all fitted means, contrasts, Wald intervals, SEs, variances, covariances, correlations, and ratios. Its Wald intervals must not be described as the primary paired bootstrap intervals. The specification is an observed-data sensitivity analysis, not a recovery of unavailable outcomes.

\subsection{Descriptive summaries and dependence}
Tables~S1--S4 report Study~1 results, and Tables~S5--S9 report Study~2 results. Table~S9 explicitly distinguishes all-measurable A2 ($N=319$), jointly measurable A1/A2 ($N=287$), and all-measurable A1 ($N=304$). The raw paired differences $+0.516488$/$+0.160013$ do not replace fitted contrasts $+0.512918$/$+0.158853$.

The sign quadrants use $\dID_k>0$ versus $\dID_k\le0$ and $\dID_j<0$ versus $\dID_j\ge0$, without qualitative task labels. Pearson/Spearman coefficients are descriptive: A1 $0.090640$/$0.026326$, A0 $0.491008$/$0.452616$, A2 $0.826727$/$0.461802$. Repeated seeds limit independent-row inference. No cluster-adjusted correlation inference is reported; nominal row-level $p$ annotations in the unchanged A2 scatter asset are not used for inference.

%% file: supplement/frozen_tables.tex
\section{Detailed numerical tables}
\label{sec:s_tables}
Tables~S1--S9 report the complete accounting, estimates, descriptive summaries, and sensitivity analyses for Study~1 and Study~2. Main Table~2 summarizes the primary scientific results.
\clearpage
\begin{landscape}
\input{supplement/tables/table_r1_accounting}
\clearpage
\input{supplement/tables/table_r2_estimates}
\clearpage
\input{supplement/tables/table_r3_sensitivity}
\clearpage
\input{supplement/tables/table_r4_joint_quadrants}
\clearpage
\input{supplement/tables/table_s2_r1_accounting}
\clearpage
\input{supplement/tables/table_s2_r2_estimates}
\clearpage
\input{supplement/tables/table_s2_r3_sensitivity}
\clearpage
\input{supplement/tables/table_s2_r4_supportive}
\clearpage
\input{supplement/tables/table_s2_r5_descriptive_joint}
\end{landscape}
\clearpage

%% file: supplement/tables/table_r1_accounting.tex
\begin{table}[H]
\centering
\small
\caption{\textbf{Confirmatory Cohort, Realization Universe, and Evaluability Accounting.} Cross-tabulation of planned versus realized cases, action-specific correspondence stages, evaluability rates, and proximate causes of missingness under Action A1 (source-coupled single-target identity-directed intervention) and Action A0 (matched global regeneration).}
\label{tab:r1_accounting}
\begin{adjustbox}{max width=\linewidth}
\begin{tabular}{lcccc}
\toprule
\textbf{Accounting Level / Metric} & \textbf{Action A1} & \textbf{Action A0} & \textbf{Total / Paired} & \textbf{Rate / Ratio} \\
\midrule
\multicolumn{5}{l}{\textit{Cohort Accrual and Realization Universe}} \\
Planned valid PRE cohort ($N$) & 120 & 120 & 120 & --- \\
Achieved valid PRE cohort ($N$) & 119 & 119 & 119 & 99.17\% \\
Prespecified stochastic seeds ($R$) & 3 & 3 & 3 & [1101, 2202, 3303] \\
Total planned POST realizations & 357 & 357 & 714 & 100.0\% \\
\midrule
\multicolumn{5}{l}{\textit{Action-Specific Correspondence Stages}} \\
Correspondence protocol & Distance-IoU & Human Triplet & --- & Independent \\
Entered structural correspondence & 357 & --- & --- & 100.0\% \\
Technical gate pass (exactly 2 faces) & --- & 308 & --- & 86.27\% \\
Human consensus accepted & --- & 293 & --- & 82.07\% \\
\textbf{Correspondence resolved} & \textbf{304} & \textbf{293} & \textbf{597} & \textbf{83.61\%} \\
Evaluability rate & 85.15\% & 82.07\% & 83.61\% & --- \\
\midrule
\multicolumn{5}{l}{\textit{Missingness Breakdown by Mechanism}} \\
Structural assignment failure (DIoU) & 53 & --- & 53 & 14.85\% \\
\quad \textit{Insufficient faces ($<2$ detected)} & 35 & --- & 35 & 9.80\% \\
\quad \textit{Tracking collision (same POST face)} & 18 & --- & 18 & 5.04\% \\
Technical missingness (face count $\neq 2$) & --- & 49 & 49 & 13.73\% \\
\quad \textit{Single face detected ($=1$)} & --- & 9 & 9 & 2.52\% \\
\quad \textit{Excess faces detected ($>2$)} & --- & 40 & 40 & 11.20\% \\
Human correspondence unresolved & --- & 15 & 15 & 4.20\% \\
Total non-evaluable realizations & 53 & 64 & 117 & 16.39\% \\
\midrule
\multicolumn{5}{l}{\textit{Case and Pair Replicate Coverage}} \\
Cases with $\ge 1$ evaluable replicate & 116 / 119 & 113 / 119 & 110 / 119 & 92.44\% \\
Cases with complete 3-seed replicates & 80 / 119 & 78 / 119 & 57 / 119 & 47.90\% \\
Case-seed pairs evaluable under both actions & --- & --- & 257 / 357 & 72.0\% \\
\bottomrule
\end{tabular}
\end{adjustbox}
\end{table}

%% file: supplement/tables/table_r2_estimates.tex
\begin{table}[H]
\centering
\small
\caption{\textbf{Primary and Secondary Inferential Estimates and Model-Based Contrasts.} Parameter estimates from the locked REML mixed-effects model across $N=119$ cases ($n=597$ evaluable realizations) with standard errors and 95\% confidence intervals computed via case-cluster bootstrap ($B=5\,000$ resamples).}
\label{tab:r2_estimates}
\begin{adjustbox}{max width=\linewidth}
\begin{tabular}{llcccc}
\toprule
\textbf{Endpoint} & \textbf{Parameter / Estimand} & \textbf{Notation} & \textbf{Point Est.} & \textbf{Bootstrap SE} & \textbf{95\% Bootstrap CI} \\
\midrule
\multicolumn{6}{l}{\textit{Primary Protected-Person Endpoint ($\Delta \mathrm{ID}_j = \mathrm{ID}_{j,\mathrm{post}} - \mathrm{ID}_{j,\mathrm{pre}}$)}} \\
& A1 Marginal Mean & $\theta_{\mathrm{A1},j}$ & \textbf{$-0.567904$} & $0.015604$ & [$-0.597297$, $-0.536832$] \\
& A0 Marginal Mean & $\theta_{\mathrm{A0},j}$ & $-0.010801$ & $0.010925$ & [$-0.030899$, $0.011336$] \\
& \textbf{A1 $-$ A0 Process Contrast} & $\theta_{D,j}$ & \textbf{$-0.557103$} & $0.013219$ & [$\mathbf{-0.582850}$, $\mathbf{-0.530484}$] \\
\cmidrule(lr){2-6}
& Between-Case Variance (A1) & $\hat{\tau}_{\mathrm{A1}}^2$ & $0.028059$ & --- & --- \\
& Between-Case Variance (A0) & $\hat{\tau}_{\mathrm{A0}}^2$ & $0.008743$ & --- & --- \\
& Case-Level Covariance & $\hat{\tau}_{\mathrm{A0},\mathrm{A1}}$ & $0.011380$ & --- & Case Corr. $\rho = 0.7266$ \\
& Within-Seed Residual Variance (A1) & $\hat{\sigma}_{\mathrm{A1}}^2$ & $0.002326$ & --- & --- \\
& Within-Seed Residual Variance (A0) & $\hat{\sigma}_{\mathrm{A0}}^2$ & $0.009076$ & --- & Ratio $\hat{\sigma}_{\mathrm{A0}}^2 / \hat{\sigma}_{\mathrm{A1}}^2 = 3.902$ \\
\midrule
\multicolumn{6}{l}{\textit{Secondary Target-Person Endpoint ($\Delta \mathrm{ID}_k = \mathrm{ID}_{k,\mathrm{post}} - \mathrm{ID}_{k,\mathrm{pre}}$)}} \\
& A1 Marginal Mean & $\theta_{\mathrm{A1},k}$ & \textbf{$-0.236904$} & $0.014251$ & [$-0.265309$, $-0.209629$] \\
& A0 Marginal Mean & $\theta_{\mathrm{A0},k}$ & $-0.021632$ & $0.009013$ & [$-0.038848$, $-0.003419$] \\
& \textbf{A1 $-$ A0 Process Contrast} & $\theta_{D,k}$ & \textbf{$-0.215272$} & $0.015603$ & [$\mathbf{-0.245432}$, $\mathbf{-0.184669}$] \\
\cmidrule(lr){2-6}
& Between-Case Variance (A1) & $\hat{\tau}_{\mathrm{A1}}^2$ & $0.014636$ & --- & --- \\
& Between-Case Variance (A0) & $\hat{\tau}_{\mathrm{A0}}^2$ & $0.004624$ & --- & --- \\
& Case-Level Covariance & $\hat{\tau}_{\mathrm{A0},\mathrm{A1}}$ & $0.002716$ & --- & Case Corr. $\rho = 0.3301$ \\
& Within-Seed Residual Variance (A1) & $\hat{\sigma}_{\mathrm{A1}}^2$ & $0.024205$ & --- & --- \\
& Within-Seed Residual Variance (A0) & $\hat{\sigma}_{\mathrm{A0}}^2$ & $0.010691$ & --- & Ratio $\hat{\sigma}_{\mathrm{A0}}^2 / \hat{\sigma}_{\mathrm{A1}}^2 = 0.442$ \\
\bottomrule
\end{tabular}
\end{adjustbox}
\end{table}

%% file: supplement/tables/table_r3_sensitivity.tex
\begin{table}[H]
\centering
\small
\caption{\textbf{Prespecified Sensitivity Analyses for Protected and Target Process Contrasts.} Comparison of estimated paired contrast statistics across varying evaluability and design subsets: available-replicate paired cases (S1), complete 3-seed replicate cases (S2), and seed-specific stratified subsets (S3).}
\label{tab:r3_sensitivity}
\begin{adjustbox}{max width=\linewidth}
\begin{tabular}{llccccc}
\toprule
\textbf{Sensitivity Analysis} & \textbf{Subset Criteria} & \textbf{$N$ Cases} & \textbf{Contrast Mean} & \textbf{Std. Error} & \textbf{95\% Conf. Interval} & \textbf{Median} \\
\midrule
\multicolumn{7}{l}{\textit{Protected-Person Contrast ($\theta_{D,j}$: A1 vs.\ A0 Difference in $\Delta \mathrm{ID}_j$)}} \\
Full REML Model & All evaluable ($n=597$) & 119 & $-0.557103$ & $0.013219$ & [$-0.582850$, $-0.530484$] & --- \\
S1: Available Paired & $\ge 1$ replicate each & 110 & $-0.561320$ & $0.013426$ & [$-0.586923$, $-0.533422$] & $-0.591209$ \\
S2: Complete Replicates & All 3 seeds each & 57 & $-0.586572$ & $0.013332$ & [$-0.611899$, $-0.560107$] & $-0.594948$ \\
S3: Seed 1101 Stratum & Seed 1101 paired & 82 & $-0.570254$ & $0.016683$ & [$-0.602640$, $-0.536939$] & $-0.584286$ \\
S3: Seed 2202 Stratum & Seed 2202 paired & 83 & $-0.572318$ & $0.018817$ & [$-0.606532$, $-0.533361$] & $-0.616727$ \\
S3: Seed 3303 Stratum & Seed 3303 paired & 92 & $-0.564490$ & $0.013706$ & [$-0.591183$, $-0.536441$] & $-0.576498$ \\
\midrule
\multicolumn{7}{l}{\textit{Secondary Target-Person Contrast ($\theta_{D,k}$: A1 vs.\ A0 Difference in $\Delta \mathrm{ID}_k$)}} \\
Full REML Model & All evaluable ($n=597$) & 119 & $-0.215272$ & $0.015603$ & [$-0.245432$, $-0.184669$] & --- \\
S1: Available Paired & $\ge 1$ replicate each & 110 & $-0.214189$ & $0.016005$ & [$-0.244378$, $-0.181856$] & $-0.207577$ \\
S2: Complete Replicates & All 3 seeds each & 57 & $-0.212455$ & $0.023530$ & [$-0.257989$, $-0.164746$] & $-0.219979$ \\
S3: Seed 1101 Stratum & Seed 1101 paired & 82 & $-0.195704$ & $0.024445$ & [$-0.242807$, $-0.147015$] & $-0.169762$ \\
S3: Seed 2202 Stratum & Seed 2202 paired & 83 & $-0.200006$ & $0.025303$ & [$-0.249350$, $-0.151352$] & $-0.178222$ \\
S3: Seed 3303 Stratum & Seed 3303 paired & 92 & $-0.228801$ & $0.021254$ & [$-0.269674$, $-0.187754$] & $-0.228308$ \\
\bottomrule
\end{tabular}
\end{adjustbox}
\end{table}

%% file: supplement/tables/table_r4_joint_quadrants.tex
\begin{table}[H]
\centering
\small
\caption{\textbf{Exploratory Joint Target/Protected Outcome Distribution and Association.} Cross-tabulation of realized transitions into joint sign quadrants defined by the joint signs of $(\Delta \mathrm{ID}_k, \Delta \mathrm{ID}_j)$, alongside linear (Pearson $r$) and rank (Spearman $\rho$) correlation coefficients under Actions A1 and A0. Correlation coefficients are exploratory descriptive summaries; cluster-adjusted correlation inference is not reported.}
\label{tab:r4_joint_quadrants}
\begin{adjustbox}{max width=\linewidth}
\begin{tabular}{llcccc}
\toprule
\textbf{Quadrant / Sign Partition} & \textbf{Condition} & \multicolumn{2}{c}{\textbf{Action A1 ($n=304$)}} & \multicolumn{2}{c}{\textbf{Action A0 ($n=293$)}} \\
\cmidrule(lr){3-4} \cmidrule(lr){5-6}
& & \textbf{Count} & \textbf{Percentage} & \textbf{Count} & \textbf{Percentage} \\
\midrule
\multicolumn{6}{l}{\textit{Joint Sign Quadrants}} \\
Quadrant 1 (Q1) & $\Delta \mathrm{ID}_k > 0 \;\land\; \Delta \mathrm{ID}_j < 0$ & 29 & 9.54\% & 45 & 15.36\% \\
Quadrant 2 (Q2) & $\Delta \mathrm{ID}_k > 0 \;\land\; \Delta \mathrm{ID}_j \ge 0$ & 0 & 0.00\% & 77 & 26.28\% \\
Quadrant 3 (Q3) & $\Delta \mathrm{ID}_k \le 0 \;\land\; \Delta \mathrm{ID}_j < 0$ & 275 & 90.46\% & 118 & 40.27\% \\
Quadrant 4 (Q4) & $\Delta \mathrm{ID}_k \le 0 \;\land\; \Delta \mathrm{ID}_j \ge 0$ & 0 & 0.00\% & 53 & 18.09\% \\
\midrule
\multicolumn{6}{l}{\textit{Marginal Sign Totals}} \\
Target positive & $\Delta \mathrm{ID}_k > 0$ & 29 & 9.54\% & 122 & 41.64\% \\
Target non-positive & $\Delta \mathrm{ID}_k \le 0$ & 275 & 90.46\% & 171 & 58.36\% \\
Protected negative & $\Delta \mathrm{ID}_j < 0$ & 304 & 100.00\% & 163 & 55.63\% \\
Protected non-negative & $\Delta \mathrm{ID}_j \ge 0$ & 0 & 0.00\% & 130 & 44.37\% \\
\midrule
\multicolumn{6}{l}{\textit{Bivariate Association Statistics}} \\
Pearson correlation ($r$) & Linear dependency & \multicolumn{2}{c}{$r = 0.090640$} & \multicolumn{2}{c}{$r = 0.491008$} \\
Spearman rank correlation ($\rho$) & Monotonic association & \multicolumn{2}{c}{$\rho = 0.026326$} & \multicolumn{2}{c}{$\rho = 0.452616$} \\
\bottomrule
\end{tabular}
\end{adjustbox}
\end{table}

%% file: supplement/tables/table_s2_r1_accounting.tex
\begin{table}[H]
\centering
\small
\caption{\textbf{Study-2 Realization Universe and Measurement Availability Accounting.} Cross-tabulation of evaluability status across the full prospective universe of $N=119$ cases evaluated across three prespecified stochastic seeds ($357$ planned realizations per action).}
\label{tab:s2_r1_accounting}
\begin{adjustbox}{max width=\linewidth}
\begin{tabular}{lcccc}
\toprule
\textbf{Accounting Level / Classification} & \textbf{Action A1} & \textbf{Action A2} & \textbf{Combined / Paired} & \textbf{Rate (\%)} \\
\midrule
\multicolumn{5}{l}{\textit{Cohort and Planned Realization Universe}} \\
Accrued valid PRE cases ($N$) & 119 & 119 & 119 & 100.0\% \\
Prespecified stochastic seeds ($R$) & 3 & 3 & 3 & [1101, 2202, 3303] \\
Total planned POST realizations & 357 & 357 & 357 pairs & 100.0\% \\
\midrule
\multicolumn{5}{l}{\textit{Action-Specific Evaluability and Measurement Status}} \\
Evaluated under DIoU structural correspondence & 357 & 357 & --- & 100.0\% \\
Structural correspondence resolved (ASSIGNED) & 304 & 319 & --- & --- \\
\textbf{Identity measurable realizations} & \textbf{304} & \textbf{319} & --- & --- \\
Marginal evaluability rate & 85.15\% & 89.36\% & --- & --- \\
Structural missingness (collision / $<2$ faces) & 53 & 38 & --- & --- \\
Marginal missingness rate & 14.85\% & 10.64\% & --- & --- \\
\midrule
\multicolumn{5}{l}{\textit{Cross-Action Joint Evaluability Partition ($357$ Planned Pairs)}} \\
\textbf{Paired measurable realizations (A1=1, A2=1)} & --- & --- & \textbf{287} & \textbf{80.39\%} \\
A1 measurable only (A1=1, A2=0) & --- & --- & 17 & 4.76\% \\
A2 measurable only (A1=0, A2=1) & --- & --- & 32 & 8.96\% \\
Neither action measurable (A1=0, A2=0) & --- & --- & 21 & 5.88\% \\
Total evaluable action realizations & 304 & 319 & 623 & 87.25\% \\
\midrule
\multicolumn{5}{l}{\textit{Case-Level Replicate Coverage}} \\
Cases with $\ge 1$ paired evaluable realization & --- & --- & 111 / 119 & 93.28\% \\
Cases with complete 3-seed paired realizations & --- & --- & 74 / 119 & 62.18\% \\
\bottomrule
\end{tabular}
\end{adjustbox}
\end{table}

%% file: supplement/tables/table_s2_r2_estimates.tex
\begin{table}[H]
\centering
\small
\caption{\textbf{Primary Protected and Key Secondary Target Paired Estimates (A2 vs.\ A1).} Model-based paired contrasts and variance components from the locked paired linear mixed-effects model ($D \sim \mathrm{Seed} + (1 \mid \mathrm{Case})$) evaluated across $287$ jointly measurable case$\times$seed pairs from $111$ distinct contributing cases with $95\%$ confidence intervals computed via case-cluster bootstrap ($B=5\,000$ resamples).}
\label{tab:s2_r2_estimates}
\begin{adjustbox}{max width=\linewidth}
\begin{tabular}{llcccc}
\toprule
\textbf{Endpoint} & \textbf{Parameter / Estimand} & \textbf{Notation} & \textbf{Point Est.} & \textbf{Bootstrap SE} & \textbf{95\% Bootstrap CI} \\
\midrule
\multicolumn{6}{l}{\textit{Primary Protected-Person Paired Contrast ($D_j = \mathrm{A2}\_\Delta \mathrm{ID}_j - \mathrm{A1}\_\Delta \mathrm{ID}_j$)}} \\
& \textbf{Equal-Seed-Weighted Paired Contrast} & $\hat{\theta}_{\mathrm{paired},j}$ & \textbf{$+0.512918$} & $0.020057$ & [\textbf{$+0.471434$}, \textbf{$+0.550259$}] \\
& Fitted Seed Marginal Mean (Seed 1101) & $\mu_{1101,j}$ & $+0.515203$ & --- & Sample Mean: $+0.522529$ \\
& Fitted Seed Marginal Mean (Seed 2202) & $\mu_{2202,j}$ & $+0.520421$ & --- & Sample Mean: $+0.520796$ \\
& Fitted Seed Marginal Mean (Seed 3303) & $\mu_{3303,j}$ & $+0.503129$ & --- & Sample Mean: $+0.506219$ \\
\cmidrule(lr){2-6}
& Between-Case Variance & $\hat{\sigma}_u^2$ & $0.031244$ & --- & --- \\
& Within-Case Residual Variance & $\hat{\sigma}_e^2$ & $0.033330$ & --- & Variance Ratio $\tau = 0.937428$ \\
& Bootstrap Convergence & $B_{\mathrm{valid}} / B$ & $5\,000 / 5\,000$ & --- & $100\%$ Success ($0$ failures) \\
\midrule
\multicolumn{6}{l}{\textit{Key Secondary Target-Person Paired Contrast ($D_k = \mathrm{A2}\_\Delta \mathrm{ID}_k - \mathrm{A1}\_\Delta \mathrm{ID}_k$)}} \\
& \textbf{Equal-Seed-Weighted Paired Contrast} & $\hat{\theta}_{\mathrm{paired},k}$ & \textbf{$+0.158853$} & $0.020346$ & [\textbf{$+0.118903$}, \textbf{$+0.197523$}] \\
& Fitted Seed Marginal Mean (Seed 1101) & $\mu_{1101,k}$ & $+0.183288$ & --- & Sample Mean: $+0.185371$ \\
& Fitted Seed Marginal Mean (Seed 2202) & $\mu_{2202,k}$ & $+0.136479$ & --- & Sample Mean: $+0.135895$ \\
& Fitted Seed Marginal Mean (Seed 3303) & $\mu_{3303,k}$ & $+0.156792$ & --- & Sample Mean: $+0.159552$ \\
\cmidrule(lr){2-6}
& Between-Case Variance & $\hat{\sigma}_u^2$ & $0.023735$ & --- & --- \\
& Within-Case Residual Variance & $\hat{\sigma}_e^2$ & $0.056765$ & --- & Variance Ratio $\tau = 0.418129$ \\
& Bootstrap Convergence & $B_{\mathrm{valid}} / B$ & $5\,000 / 5\,000$ & --- & $100\%$ Success ($0$ failures) \\
\bottomrule
\end{tabular}
\end{adjustbox}
\end{table}

%% file: supplement/tables/table_s2_r3_sensitivity.tex
\begin{table}[H]
\centering
\small
\caption{\textbf{Prespecified Sensitivity Analyses for Protected and Target Paired Contrasts.} Robustness of the paired $A2 - A1$ contrast across complete 3-seed replicate cases (S1) and seed-stratified strata (S2). Confidence intervals computed via case-resampled bootstrap ($B=5\,000$).}
\label{tab:s2_r3_sensitivity}
\begin{adjustbox}{max width=\linewidth}
\begin{tabular}{llcccc}
\toprule
\textbf{Sensitivity Analysis} & \textbf{Subset Criteria} & \textbf{$N$ Cases} & \textbf{$N$ Pairs} & \textbf{Contrast Mean} & \textbf{95\% Conf. Interval} \\
\midrule
\multicolumn{6}{l}{\textit{Protected-Person Contrast ($D_j = \mathrm{A2}\_\Delta \mathrm{ID}_j - \mathrm{A1}\_\Delta \mathrm{ID}_j$)}} \\
Primary Paired REML & All paired evaluable & 111 & 287 & $+0.512918$ & [$+0.471434$, $+0.550259$] \\
S1: Complete 3-Seed Cases & Valid pairs at all 3 seeds & 74 & 222 & $+0.526683$ & [$+0.479948$, $+0.571326$] \\
S2: Seed 1101 Stratum & Seed 1101 paired subset & 94 & 94 & $+0.522529$ & [$+0.471128$, $+0.572450$] \\
S2: Seed 2202 Stratum & Seed 2202 paired subset & 97 & 97 & $+0.520796$ & [$+0.470800$, $+0.569094$] \\
S2: Seed 3303 Stratum & Seed 3303 paired subset & 96 & 96 & $+0.506219$ & [$+0.452226$, $+0.554927$] \\
\midrule
\multicolumn{6}{l}{\textit{Secondary Target-Person Contrast ($D_k = \mathrm{A2}\_\Delta \mathrm{ID}_k - \mathrm{A1}\_\Delta \mathrm{ID}_k$)}} \\
Primary Paired REML & All paired evaluable & 111 & 287 & $+0.158853$ & [$+0.118903$, $+0.197523$] \\
S1: Complete 3-Seed Cases & Valid pairs at all 3 seeds & 74 & 222 & $+0.169166$ & [$+0.118506$, $+0.217843$] \\
S2: Seed 1101 Stratum & Seed 1101 paired subset & 94 & 94 & $+0.185371$ & [$+0.132393$, $+0.234278$] \\
S2: Seed 2202 Stratum & Seed 2202 paired subset & 97 & 97 & $+0.135895$ & [$+0.075700$, $+0.193176$] \\
S2: Seed 3303 Stratum & Seed 3303 paired subset & 96 & 96 & $+0.159552$ & [$+0.099867$, $+0.215647$] \\
\bottomrule
\end{tabular}
\end{adjustbox}
\end{table}

%% file: supplement/tables/table_s2_r4_supportive.tex
\begin{table}[H]
\centering
\small
\caption{\textbf{Supportive All-Available Linear Mixed-Effects Model Estimates.} Model-conditional parameter estimates from the locked supportive mixed model ($\Delta \mathrm{ID} \sim \mathrm{Action} \times \mathrm{Seed} + (1 + \mathrm{Action} \mid \mathrm{Case}) + (1 \mid \mathrm{Case:Seed})$) evaluated across all $N=623$ evaluable action realizations ($304$ under A1, $319$ under A2) using profile REML. These results are supportive and observed-data-conditional; they are directionally consistent with the primary paired analysis and do not replace it.}
\label{tab:s2_r4_supportive}
\begin{adjustbox}{max width=\linewidth}
\begin{tabular}{llcccc}
\toprule
\textbf{Endpoint} & \textbf{Parameter / Estimand} & \textbf{Notation} & \textbf{Point Est.} & \textbf{Std. Error} & \textbf{95\% Wald CI} \\
\midrule
\multicolumn{6}{l}{\textit{Supportive Protected-Person Outcome ($\Delta \mathrm{ID}_j$, $N=623$ Evaluable Realizations)}} \\
& A1 Equal-Seed Marginal Mean & $\mu_{\mathrm{A1},j}$ & $-0.568657$ & --- & --- \\
& A2 Equal-Seed Marginal Mean & $\mu_{\mathrm{A2},j}$ & $-0.070088$ & --- & --- \\
& \textbf{Marginal Action Contrast (A2 $-$ A1)} & $\theta_{\mathrm{S3},j}$ & \textbf{$+0.498569$} & $0.020300$ & [\textbf{$+0.458782$}, \textbf{$+0.538357$}] \\
\cmidrule(lr){2-6}
& Case:Seed Random Variance & $\hat{\sigma}_{cs}^2$ & $0.000860$ & --- & Non-singular ($>10^{-6}$) \\
& Between-Case Variance (A1) & $\hat{\tau}_{\mathrm{A1}}^2$ & $0.027820$ & --- & --- \\
& Between-Case Variance (A2) & $\hat{\tau}_{\mathrm{A2}}^2$ & $0.009089$ & --- & --- \\
& Case-Level Covariance & $\hat{\tau}_{\mathrm{A1},\mathrm{A2}}$ & $0.001198$ & --- & Correlation $\rho = 0.0754$ \\
& Within-Seed Residual Variance (A1) & $\hat{\sigma}_{\mathrm{A1}}^2$ & $0.001472$ & --- & --- \\
& Within-Seed Residual Variance (A2) & $\hat{\sigma}_{\mathrm{A2}}^2$ & $0.033554$ & --- & Ratio $\hat{\sigma}_{\mathrm{A2}}^2 / \hat{\sigma}_{\mathrm{A1}}^2 = 22.79$ \\
\midrule
\multicolumn{6}{l}{\textit{Supportive Target-Person Outcome ($\Delta \mathrm{ID}_k$, $N=623$ Evaluable Realizations)}} \\
& A1 Equal-Seed Marginal Mean & $\mu_{\mathrm{A1},k}$ & $-0.237109$ & --- & --- \\
& A2 Equal-Seed Marginal Mean & $\mu_{\mathrm{A2},k}$ & $-0.079556$ & --- & --- \\
& \textbf{Marginal Action Contrast (A2 $-$ A1)} & $\theta_{\mathrm{S3},k}$ & \textbf{$+0.157553$} & $0.019354$ & [\textbf{$+0.119619$}, \textbf{$+0.195486$}] \\
\cmidrule(lr){2-6}
& Case:Seed Random Variance & $\hat{\sigma}_{cs}^2$ & $0.001964$ & --- & Non-singular ($>10^{-6}$) \\
& Between-Case Variance (A1) & $\hat{\tau}_{\mathrm{A1}}^2$ & $0.014569$ & --- & --- \\
& Between-Case Variance (A2) & $\hat{\tau}_{\mathrm{A2}}^2$ & $0.008242$ & --- & --- \\
& Case-Level Covariance & $\hat{\tau}_{\mathrm{A1},\mathrm{A2}}$ & $0.000827$ & --- & Correlation $\rho = 0.0755$ \\
& Within-Seed Residual Variance (A1) & $\hat{\sigma}_{\mathrm{A1}}^2$ & $0.022278$ & --- & --- \\
& Within-Seed Residual Variance (A2) & $\hat{\sigma}_{\mathrm{A2}}^2$ & $0.035526$ & --- & Ratio $\hat{\sigma}_{\mathrm{A2}}^2 / \hat{\sigma}_{\mathrm{A1}}^2 = 1.595$ \\
\bottomrule
\end{tabular}
\end{adjustbox}
\end{table}

%% file: supplement/tables/table_s2_r5_descriptive_joint.tex
\begin{table}[H]
\centering
\small
\setlength{\tabcolsep}{3pt}
\caption{\textbf{Descriptive Outcome Distributions and Joint Quadrant Analysis.} Continuous identity transition distributions ($\Delta \mathrm{ID}$) across explicitly labeled populations: all-measurable A2 realizations ($N=319$), jointly measurable paired realizations ($N=287$), and all-measurable A1 realizations ($N=304$). Quadrants partition the joint space of target and protected transitions without qualitative labels. Correlation coefficients are exploratory descriptive summaries.}
\label{tab:s2_r5_descriptive_joint}
\begin{adjustbox}{max width=\linewidth}
\begin{tabular}{lccccccc}
\toprule
\shortstack{\textbf{Population /}\\\textbf{Endpoint}} & \textbf{$N$} & \textbf{Mean} & \textbf{SD} & \textbf{Median} & \textbf{IQR [Q1, Q3]} & \textbf{Min} & \textbf{Max} \\
\midrule
\multicolumn{8}{l}{\textit{Panel A: Action A2 Absolute Distributions ($N=319$ All-Measurable A2 Realizations)}} \\
Protected Person ($\Delta \mathrm{ID}_j$) & $N=319$ & $-0.070549$ & $0.208261$ & $-0.000015$ & [$-0.064465$, $+0.035160$] & $-0.773612$ & $+0.342372$ \\
Target Person ($\Delta \mathrm{ID}_k$) & $N=319$ & $-0.080126$ & $0.213588$ & $-0.015024$ & [$-0.077919$, $+0.029863$] & $-0.804875$ & $+0.301640$ \\
\midrule
\multicolumn{8}{l}{\textit{Panel B: Direct Paired Comparison on Jointly Measurable Population ($N=287$ Identical Denominator)}} \\
A1 Protected ($\Delta \mathrm{ID}_j$) & $N=287$ & $-0.589143$ & $0.155775$ & $-0.621328$ & [$-0.697440$, $-0.527795$] & $-0.889022$ & $-0.036281$ \\
A2 Protected ($\Delta \mathrm{ID}_j$) & $N=287$ & $-0.072655$ & $0.211454$ & $+0.000230$ & [$-0.063106$, $+0.035160$] & $-0.773612$ & $+0.165424$ \\
\cmidrule(lr){2-8}
A1 Target ($\Delta \mathrm{ID}_k$) & $N=287$ & $-0.241357$ & $0.199355$ & $-0.207624$ & [$-0.399220$, $-0.084950$] & $-0.703184$ & $+0.306814$ \\
A2 Target ($\Delta \mathrm{ID}_k$) & $N=287$ & $-0.081344$ & $0.215198$ & $-0.016084$ & [$-0.077919$, $+0.029863$] & $-0.804875$ & $+0.301640$ \\
\midrule
\multicolumn{8}{l}{\textit{Panel C: Reference Single-Action A1 Population ($N=304$ All-Measurable A1 Realizations)}} \\
Protected Person ($\Delta \mathrm{ID}_j$) & $N=304$ & $-0.578661$ & $0.164615$ & $-0.616876$ & [$-0.692620$, $-0.508078$] & $-0.889022$ & $-0.036281$ \\
Target Person ($\Delta \mathrm{ID}_k$) & $N=304$ & $-0.239163$ & $0.197195$ & $-0.207292$ & [$-0.395470$, $-0.085080$] & $-0.703184$ & $+0.306814$ \\
\midrule
\multicolumn{8}{l}{\textit{Panel D: Action A2 Joint Outcome Quadrant Occupancy ($N=319$ All-Measurable Realizations)}} \\
\multicolumn{3}{l}{Quadrant 1 ($\Delta \mathrm{ID}_k > 0$ and $\Delta \mathrm{ID}_j < 0$):} & \multicolumn{2}{c}{$46$ realizations ($14.42\%$)} & \multicolumn{3}{l}{Pearson correlation: $r = 0.826727$} \\
\multicolumn{3}{l}{Quadrant 2 ($\Delta \mathrm{ID}_k > 0$ and $\Delta \mathrm{ID}_j \ge 0$):} & \multicolumn{2}{c}{$91$ realizations ($28.53\%$)} & \multicolumn{3}{l}{Spearman rank correlation: $\rho = 0.461802$} \\
\multicolumn{3}{l}{Quadrant 3 ($\Delta \mathrm{ID}_k \le 0$ and $\Delta \mathrm{ID}_j < 0$):} & \multicolumn{2}{c}{$114$ realizations ($35.74\%$)} & \multicolumn{3}{l}{Quadrant criteria: sign of continuous $\Delta \mathrm{ID}$ transition} \\
\multicolumn{3}{l}{Quadrant 4 ($\Delta \mathrm{ID}_k \le 0$ and $\Delta \mathrm{ID}_j \ge 0$):} & \multicolumn{2}{c}{$68$ realizations ($21.32\%$)} & \multicolumn{3}{l}{No automated qualitative success/failure interpretation} \\
\bottomrule
\end{tabular}
\end{adjustbox}
\end{table}

%% file: supplement/independent_validation.tex
\section{Independent metric and human validation records}
\label{sec:s_independent}
The primary Study~2 result was frozen before prospective validation design and execution. The specifications were fixed before validation execution. Table~S10 summarizes the validation estimates and distinguishes independent measurement from within-evaluator sensitivity. The protocol classifies all directions as CONCORDANT and the integrated result as OVERALL\_CONCORDANT. These are directional classifications, not independent experimental replications.

\input{supplement/tables/validation_summary}

\subsection{V1: metric implementation and sign}
The independent identity-focused evaluator is ID-Sim \citep{chae2026idsim}, using frozen checkpoint \path{dinov3_vitl16_cls_patch} in \texttt{cls} mode. Lower distance $d$ means greater consistency. The frozen convention is
\begin{equation}
\Delta\ID_{\mathrm{IDSim},p,a}=d(\mathrm{REF}_p,\PRE_p)-d(\mathrm{REF}_p,\POST_{p,a}),\qquad
D_{\mathrm{IDSim},p}=\Delta\ID_{\mathrm{IDSim},p,\Atwo}-\Delta\ID_{\mathrm{IDSim},p,\Aone}.
\end{equation}
Thus positive $D$ favors A2; a lower-distance metric is not inadvertently interpreted with a cosine sign. The identical 287 paired rows and 111 cases enter the paired seed/random-case model, with 5,000 case-cluster bootstrap resamples and seed 20260927. Estimates and confidence intervals are transcribed in Table~S10. They provide agreement on a distinct representation of the same outputs, not comparable magnitudes across feature spaces or a new full-study replication.

\subsection{V2: selection, instructions, and blinding}
The prespecified human-validation protocol selected 80 unique cases from the 111 contributing paired cases, stratified into 40 Target Person~1 and 40 Target Person~2 cases. The lexicographically smallest hash selected one jointly measurable seed per case in the frozen private blinding manifest. Both person roles used that selected case--seed realization. Each of 160 packets presented the role's reference beside A1/A2 candidates, with 80 A1-left and 80 A2-left orientations. Five final primary raters made independent judgments, yielding 800 responses. The documented final panel was an outcome-blind administrative amendment locked before decoding.

The instruction scope asks which candidate better matches the displayed reference facial identity. The four standardized response options are: Candidate LEFT better matches identity; Candidate RIGHT better matches identity; No clear difference; Unable to judge. Raters were blinded to regime labels, case IDs, model architecture, and biometric scores. Access to the complete operational protocol is subject to the data-availability and privacy constraints stated in Section~S7.

At least three of five judgeable responses are required. ``Unable to judge'' is excluded from the denominator; ``No clear difference'' is judgeable and contributes zero:
\begin{equation}
\mathcal H=\frac{N_{\mathrm{A2}}-N_{\mathrm{A1}}}{N_{\mathrm{judgeable}}},\qquad -1\le\mathcal H\le1.
\end{equation}
Mean protected preference is primary, target preference secondary. Case-cluster percentile intervals use 5,000 resamples and seed 20260928, conditional on the fixed final panel. A packet with fewer than three judgeable responses is non-estimable and is not assigned an artificial zero.

\subsection{V2: complete response and reliability summaries}
All 80 protected packets were evaluable (100\%, zero non-estimable). Of 400 ratings, 374 were judgeable (93.5\%) and 26 unable (6.5\%). There were 370 A2 choices (92.5\% of all; 98.93\% of judgeable), zero A1 (0.0\%), and four no-clear-difference responses (1.0\%). All 80 packets had $\mathcal H>0$, median 1.0, IQR 0.0. Mean preference was $+0.986667$, CI $[0.965000,1.000000]$, bootstrap SE $0.009457$.

For the target, 77 of 80 packets were evaluable (96.25\%); three were non-estimable with fewer than three judgeable ratings. Of 400 ratings, 369 were judgeable (92.25\%) and 31 unable (7.75\%). Choices were 184 A2 (46.0\% of all; 49.86\% of judgeable), 34 A1 (8.5\%; 9.21\%), and 151 no clear difference (37.75\%; 40.92\%). Among eligible packets, 59 had $\mathcal H>0$, eight $\mathcal H=0$, and ten $\mathcal H<0$, median 0.4 and IQR 0.6. Mean preference was $+0.401948$, CI $[0.304816,0.498673]$, bootstrap SE $0.049294$.

The fixed panel's mean pairwise raw agreement was 67.75\%, range 55.6\%--82.5\%; multi-category Fleiss' $\kappa=0.5308$. Leave-one-rater-out four-rater subsets gave protected means $+0.9844$ to $+0.9906$ and target means $+0.3258$ to $+0.4470$. Each omission retained the positive direction. These checks assess the influence of this panel's members. The confidence intervals are case-level conditional on the panel, not uncertainty over a larger rater population. Ethics/consent documentation remains pending institutional determination, as stated in Section~S7.

%% file: supplement/tables/validation_summary.tex
\begin{table}[H]
\centering\small
\setlength{\tabcolsep}{4pt}
\caption{Validation summary. The protocol classifications are CONCORDANT for V1/V2/V3 and OVERALL\_CONCORDANT overall. Independent metric/human measurements and supportive within-\buffalo\ nuisance sensitivity have distinct evidential roles. Intervals are 95\% case-cluster bootstrap intervals; scales are not interchangeable.}
\label{tab:s_validation}
\begin{tabularx}{\linewidth}{p{0.18\linewidth}Yp{0.27\linewidth}p{0.27\linewidth}}
\toprule
Axis & Evidential role & Protected: estimate [95\% CI] & Target: estimate [95\% CI]\\\midrule
Primary Study~2 & \buffalo\ paired anchor & $+0.512918$\newline $[0.471434,0.550259]$ & $+0.158853$\newline $[0.118903,0.197523]$\\\addlinespace
V1: ID-Sim & Independent metric concordance & $+0.231351$\newline $[0.207440,0.254732]$ & $+0.072272$\newline $[0.048170,0.097237]$\\\addlinespace
V2: Human ($N=5$ raters) & Independent human concordance & $+0.986667$\newline $[0.965000,1.000000]$ & $+0.401948$\newline $[0.304816,0.498673]$\\\addlinespace
V3: Core & Supportive within-\buffalo\ sensitivity & $+0.511637$\newline $[0.471271,0.548896]$ & $+0.158524$\newline $[0.122081,0.192932]$\\\addlinespace
V3: Pose & Supportive within-\buffalo\ sensitivity & $+0.511421$\newline $[0.471351,0.550273]$ & $+0.158543$\newline $[0.123847,0.192405]$\\
\bottomrule
\end{tabularx}
\end{table}

%% file: supplement/nuisance.tex
\section{V3 nuisance sensitivity}
\label{sec:s_v3}
V3 is supportive analysis within the original \buffalo\ outcome. Its core specification is fitted separately for protected $j$ and target $k$:
\begin{align}
D_{\mathrm{Buffalo},p}\sim {}&\mathrm{Seed}+z(D_{\log\mathrm{area},p})+z(D_{\log\mathrm{sharpness},p})\nonumber\\
&+z(D_{\mathrm{det\ confidence},p})+(1\mid\mathrm{Case}).
\end{align}
Metrics are $q_{\mathrm{size}}=\ln(\mathrm{face\ area\ ratio})$, $q_{\mathrm{sharp}}=\ln(1+\mathrm{Var}(\mathrm{Laplacian}))$, and frozen SCRFD detection confidence. The pose model adds standardized differences in $|\mathrm{yaw}|$ and $|\mathrm{pitch}|$ from 6DRepNet \citep{hempel2022sixdrepnet}; roll is descriptive only. Paired differences and frozen standardization are
\begin{equation}
D_q=q_{\Atwo}-q_{\Aone},\qquad z(D_q)=\frac{D_q-\operatorname{mean}(D_q)}{\operatorname{sd}(D_q)}.
\end{equation}
Profile REML uses a random case intercept. The adjusted marginal contrast equally averages three fitted seed means at $z(D_q)=0$, the mean observed nuisance profile. This is not literal physical equality $D_q=0$ between actions. Percentile intervals use 5,000 case-cluster bootstrap resamples with NumPy \texttt{RandomState(20260929)}. Repeated sampled occurrences of a case form independent resampled cluster blocks.

Table~S10 reports estimates to six decimal places; Table~S11 provides the corresponding higher-precision values. The positive adjusted contrasts describe sensitivity to these measured covariates. They do not establish an independent identity evaluator, causal mechanism, elimination of confounding, or exclusion of unmeasured quality effects.

\begin{table}[H]
\centering\small
\caption{V3 nuisance-adjusted estimates and 95\% case-cluster bootstrap intervals.}
\label{tab:s_v3_precision}
\begin{tabular}{llrr}
\toprule
Model & Role & Estimate & 95\% case-cluster CI\\\midrule
Core & Protected $j$ & $+0.511636943$ & $[0.47127149,0.54889617]$\\
Core & Target $k$ & $+0.158524410$ & $[0.12208135,0.19293198]$\\
Pose-augmented & Protected $j$ & $+0.511421391$ & $[0.47135144,0.55027348]$\\
Pose-augmented & Target $k$ & $+0.158542884$ & $[0.12384655,0.19240531]$\\
\bottomrule
\end{tabular}
\end{table}

%% file: supplement/plots.tex
\section{Additional statistical figures}
\label{sec:s_plots}
The following figures complement the protected-person results in main Figure~2, the qualitative comparisons in main Figure~3, and the validation results in main Figure~4. Their captions specify the relevant denominators and inference boundaries.
\begin{figure}[H]
\centering\includegraphics[width=0.78\linewidth]{fig_r2_delta_id_k_distribution.png}
\caption{Target-person transition distributions under A1 ($n=304$) and A0 ($n=293$). Zero denotes no continuous change, not a categorical success threshold.}
\label{fig:s_target_study1}
\end{figure}
\clearpage
\begin{figure}[H]
\centering\includegraphics[width=0.87\linewidth]{fig_r4_case_level_protected_contrast.png}
\caption{Study~1 case-level protected process-contrast distribution for 110 available paired cases. The mean line ($-0.561$) and median ($-0.591$) are rounded descriptive summaries of the paired-case sensitivity, not replacements for the full fitted contrast.}
\label{fig:s_case_study1}
\end{figure}
\begin{figure}[H]
\centering\includegraphics[width=\linewidth]{fig_r3_joint_scatter_by_action.png}
\caption{Study~1 joint target/protected transitions for A1 ($n=304$) and A0 ($n=293$). Sign quadrants are descriptive; target-nonpositive/protected-negative observations dominate A1.}
\label{fig:s_joint_study1}
\end{figure}
\clearpage
\begin{figure}[H]
\centering\includegraphics[width=\linewidth]{fig_s2_r2_paired_delta_k.png}
\caption{Study~2 target A2$-$A1 paired contrast. Left: 287 paired rows and fitted equal-seed mean with 95\% case-cluster bootstrap interval. Right: seed strata of 94, 97, and 96 pairs. The contrast is relative and does not imply that A2's absolute transition is positive.}
\label{fig:s_target_study2}
\end{figure}
\clearpage
\begin{figure}[H]
\centering\includegraphics[width=0.82\linewidth]{fig_s2_r3_a2_joint_scatter.png}
\caption{A2 joint transitions across all 319 measurable outputs. Counts and percentages partition the continuous sign space without qualitative success labels. \textbf{Nominal realization-level correlation $p$ annotations in the plot do not account for repeated seeds and are not used for inference.} Coefficients are descriptive. The upper-left quadrant Q4 contains 68 observations (21.32\%), as reported in Table~S9.}
\label{fig:s_joint_study2}
\end{figure}
\clearpage

%% file: supplement/provenance_ethics.tex
\section{Ethics and data availability}
\label{sec:s_provenance}
\subsection{Ethics and human-evaluation scope}
The benchmark's reference portraits derive from its documented FFHQ, CelebA-HQ, and SFHQ sources. PRE/POST images are synthetic UniPortrait outputs. No facial images or biometric measurements of raters were collected. Scientific response records use opaque rater identifiers. Human activities comprise structurally blinded quality/correspondence adjudication and the separate V2 identity comparison task; these are not conflated. Automated identity evaluators were research measurements, not deployed surveillance or physical-person identification systems. Identity distortion and non-consensual personalization remain relevant societal considerations.

\noindent\textbf{Ethics/consent documentation is pending institutional determination and must be finalized before submission according to institutional and venue requirements.} No institutional approval or exemption is asserted.

\subsection{Reproducibility and data availability}
A curated scientific reproducibility release and repository locator remain pending. Access to reference imagery, generated outputs, model assets, and human-response records is subject to material-specific licensing, redistribution permissions, and rater privacy. Restricted imagery, non-redistributable weights, identifiable rater information, credentials, and confidential ethics files must be excluded from any public release. Access to the complete operational human-evaluation protocol is subject to the same restrictions.